\documentclass[sigconf,natbib=true]{acmart}
\usepackage{booktabs} % For formal tables
\usepackage{amsthm}
\usepackage{amsmath}
\usepackage{algorithm}
\usepackage{algorithmicx}
\usepackage{algpseudocode}
\usepackage{balance}

\usepackage[titletoc,toc,title]{appendix}

\hypersetup{draft}

\usepackage{caption}
\usepackage{graphicx}
\usepackage{amsmath}
\usepackage{booktabs}

\usepackage{amsmath,bm}
\usepackage{multirow}

\usepackage{color}
\usepackage{xcolor}
\usepackage{framed}
\usepackage{diagbox}
\usepackage{booktabs}
\usepackage{enumitem}
\usepackage{multicol}
\usepackage{textcomp}

\usepackage{subfigure}

\usepackage{multirow}
\usepackage{tabularx}

\usepackage{balance}
\copyrightyear{2022}
\acmYear{2022}
\setcopyright{acmcopyright}
\acmConference[SIGIR '22]{Proceedings of the 45th International ACM SIGIR Conference on Research and Development in Information Retrieval}{July 11--15, 2022}{Madrid, Spain}
\acmBooktitle{Proceedings of the 45th International ACM SIGIR Conference on Research and Development in Information Retrieval (SIGIR '22), July 11--15, 2022, Madrid, Spain}
\acmPrice{15.00}
\acmDOI{10.1145/3477495.3531984}
\acmISBN{978-1-4503-8732-3/22/07}

\begin{document}
\bibliographystyle{ACM-Reference-Format}

\title{Graph Adaptive Semantic Transfer for Cross-domain\\ Sentiment Classification}
\author{Kai Zhang}
\affiliation{%
 \institution{Anhui Province Key Lab. of Big Data Analysis and Application, University of S\&T of China \& State Key Laboratory of Cognitive Intelligence}
 \city{Hefei}
 \country{China}
}
\email{kkzhang0808@mail.ustc.edu.cn}

\author{Qi Liu, Zhenya Huang}
\affiliation{
 \institution{Anhui Province Key Lab. of Big Data Analysis and Application, University of S\&T of China \& State Key Laboratory of Cognitive Intelligence}
 \city{Hefei}
 \country{China}
}
\email{{qiliuql, huangzhy}@ustc.edu.cn}

\author{Mingyue Cheng}
\affiliation{%
 \institution{Anhui Province Key Lab. of Big Data Analysis and Application, University of S\&T of China \& State Key Laboratory of Cognitive Intelligence}
 \city{Hefei}
 \country{China}
}
\email{mycheng@mail.ustc.edu.cn}

\author{Kun Zhang}
\affiliation{%
 \institution{School of Computer Science and Information Engineering, Hefei University of Technology}
 \city{Hefei}
 \country{China}
}
\email{zhang1024kun@gmail.com}

\author{Mengdi Zhang, Wei Wu}
\affiliation{%
 \institution{Meituan}
 \city{Beijing}
 \country{China}
}
\email{mdzhangmd@gmail.com}
\email{wuwei19850318@gmail.com}

\author{Enhong Chen}
\authornote{Corresponding Author.}
\affiliation{%
 \institution{Anhui Province Key Lab. of Big Data Analysis and Application, University of S\&T of China}
 \city{Hefei}
 \country{China}
}
\email{cheneh@ustc.edu.cn}
\renewcommand{\shortauthors}{Zhang, et al.}

\begin{CCSXML}
<ccs2012>
<concept>
<concept_id>10002951.10003317.10003347</concept_id>
<concept_desc>Information systems~Retrieval tasks and goals</concept_desc>
<concept_significance>500</concept_significance>
</concept>
<concept>
<concept_id>10002951.10003317.10003347.10003353</concept_id>
<concept_desc>Information systems~Sentiment analysis</concept_desc>
<concept_significance>500</concept_significance>
</concept>
</ccs2012>
\end{CCSXML}

\ccsdesc[500]{Information systems~Retrieval tasks and goals}
\ccsdesc[500]{Information systems~Sentiment analysis}

\begin{abstract}
Cross-domain sentiment classification (CDSC) aims to use the transferable semantics learned from the source domain to predict the sentiment of reviews in the unlabeled target domain. Existing studies in this task attach more attention to the sequence modeling of sentences while largely ignoring the rich domain-invariant semantics embedded in graph structures (i.e., the part-of-speech tags and dependency relations). As an important aspect of exploring characteristics of language comprehension, adaptive graph representations have played an essential role in recent years. To this end, in the paper, we aim to explore the possibility of learning invariant semantic features from graph-like structures in CDSC. Specifically, we present \emph{\textbf{G}raph \textbf{A}daptive \textbf{S}emantic \textbf{T}ransfer (\textbf{GAST})} model, an adaptive syntactic graph embedding method that is able to learn domain-invariant semantics from both word sequences and syntactic graphs. More specifically, we first raise a \emph{POS-Transformer} module to extract sequential semantic features from the word sequences as well as the part-of-speech tags. Then, we design a \emph{Hybrid Graph Attention (HGAT)} module to generate syntax-based semantic features by considering the transferable dependency relations. Finally, we devise an \emph{Integrated aDaptive Strategy (IDS)} to guide the joint learning process of both modules. Extensive experiments on four public datasets indicate that GAST achieves comparable effectiveness to a range of state-of-the-art models.  
\end{abstract}

\keywords{Text Mining; Sentiment Analysis; Domain Adaptation; Graph Embedding; Web Content Analysis
}

\maketitle
{\fontsize{8pt}{8pt} \selectfont
	\textbf{ACM Reference Format:}\\
	Kai Zhang, Qi Liu, Zhenya Huang, Mingyue Cheng, Kun Zhang, Mengdi Zhang, Wei Wu, Enhong Chen. 2022. Graph Adaptive Semantic Transfer for Cross-domain Sentiment Classification. In \textit{Proceedings of the 45th International ACM SIGIR Conference on Research and Development in Information Retrieval (SIGIR ’22), July 11–15, 2022, Madrid, Spain}. ACM, New York, NY, USA, 11 pages. https://doi.org/10.1145/3477495.3531984}

\section{Introduction} 
Sentiment classification is a fundamental task in natural language processing (NLP). Over the past decades, many supervised machine learning methods such as logistic regression, support vector machines, and neural networks~\cite{pang2002thumbs,hu2004mining,liu2012sentiment,zhang2022incorporating} are applied to the task. However, due to the domain shift problem, directly using the off-the-shelf sentiment classifiers to a new domain (e.g., dataset) may lead to a significant performance drop~\cite{long2015learning}. 

To address the problem, cross-domain sentiment classification (CDSC), which refers to utilizing the valuable knowledge in the source domain to help sentiment prediction in a target domain, has been proposed and extensively studied in the last decade. In the literature, many previous methods focus on learning the shared features, i.e., common sentiment words~\cite{blitzer2007biographies,pan2010cross,chen2012marginalized}, part-of-speech tags~\cite{xia2011pos} and syntactic tree~\cite{zou2015sentiment}, with traditional machine learning methods, which are usually based on handcrafted features and fail to model the deep semantic representations across domains because the universal structures are essentially human-curated and expensive to acquire across domains.

Subsequently, with the great progress of deep neural networks in numerous NLP tasks, some scholars explore deep models to learn latent representations of domain-shared information. Most of these studies~\cite {ganin2014unsupervised,ganin2016domain} focus on extracting features from the word sequences and embed the sequential features into deep semantic representations 
through various methods, e.g., memory network~\cite{li2017end}, recurrent neural network~\cite{yu-jiang-2016-learning,cai2019multi}, attention mechanism~\cite{li2018hierarchical,zhang2019interactive}, and the large-scale pre-trained models~\cite{du2020adversarial,li2020simultaneous}. However, these studies only concentrate on modeling the domain-invariant semantics from textual word sequences, while largely discarding the exploration of adaptive graph syntactic information, i.e., the part-of-speech tags  (POS-tags) and dependency relations.

Actually, as an important aspect of exploring the characteristics of language comprehension, syntactic information exploration has made significant progress, especially being combined with graph-based models in many NLP tasks~\cite{li2012cross,wang2018recursive,huang2019syntax}. For example, in ABSA, using syntactic information to enhance the semantic representation of aspects has become the basic configuration of the SOTA model~\cite{wang2020relational,tian2021aspect}. However, current advanced methods of CDSC learn semantics only from standardized word sequences while largely ignoring those adaptive syntactic structure information. Therefore, despite their popularity, efforts to incorporate universal language structure correspondence between domains such as part-of-speech tags and dependency relations into the domain adaptation framework have been sporadic. To this end, we identify multiple advantages of using adaptive syntactic-semantic for domain adaptation.

\begin{figure}
\vspace{0.15cm}
	\centering
	\includegraphics[width=1\columnwidth]{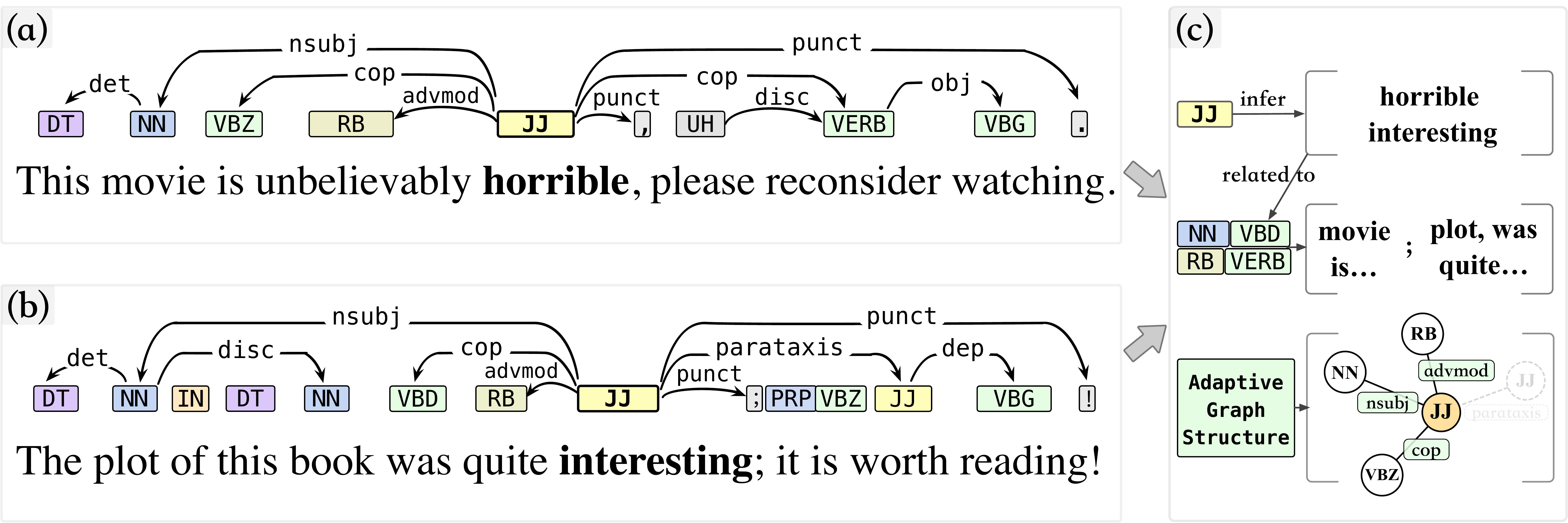}
	\vspace{-0.6cm}
	\caption{\small{
	The transferable syntactic structures\protect\footnotemark\ of two examples (i.e., (a), (b)). The colorful boxes (``DT'') and black lines (e.g., ``det'') indicate POS tags and syntactic relations, respectively. %(c) denotes the domain-adaptive information. 
	As shown in (c), the syntactic structures are similar between domains so that it is easy for human to understand the hidden knowledge behind sentences in different domains. However, those adaptive graph features are largely ignored by existing domain adaptation research. 
	}}
	\label{fig:example}
    	\vspace{-0.4cm}
\end{figure}

\textbf{First}, sentiment words play a crucial role in CDSC~\cite{xia2011pos}, while POS tags can distinguish sentiment words (e.g., \emph{``horrible''} and \emph{``interesting''} in Figure~\ref{fig:example}) via the POS tag ``JJ'' in a natural way, i.e., the ``JJ'' label means the word is an adjective. Unfortunately, recent studies only explore word semantics from the pre-trained word embeddings, which may not be sufficient to identify sentiment words in the CSDC task. 
\textbf{Second}, the sentiment polarity of reviews is largely influenced by the sentiment word's neighbors, whether they are in-domain or across-domain. As Figure~\ref{fig:example} (b) shows, the neighbor \emph{``quite''} is more important than non-adjacent words (e.g.,\emph{``the''} and \emph{``book''}) for the sentiment word \emph{``interesting''}. Meanwhile, different neighbors' syntactic relations also have different influences for each word. For instance, the neighbor word \emph{``quite''} also plays a more critical role than neighbor \emph{``was''} for sentiment word \emph{``interesting''}. 
Thus, existing methods that solely rely on sequential relations may lead to sub-optimal sentiment prediction in CDSC. \textbf{Third}, as shown in Figure~\ref{fig:example} (c), the syntactic graph structures of sentences in different domains are remarkably similar, which means that the syntactic rules are domain-invariant and can be naturally transferred across domains. However, the exploration of these crucial features is still neglected, and how to train a graph adaptive semantic transferable model for CDSC has not been fully considered.

\renewcommand{\thefootnote}{\fnsymbol{footnote}}
\footnotetext{The syntactic structure of the sentences are constructed by the Stanford CoreNLP toolkit: https://stanfordnlp.github.io/CoreNLP/.}

Following the above intuitions, in this paper, we propose a \emph{\textbf{G}raph \textbf{A}daptive \textbf{S}emantic \textbf{T}ransfer (\textbf{GAST})} model, which aims to learn textual semantics and graph adaptive semantics for cross-domain sentiment classification. Generally, GAST improves the semantic representation and transferable knowledge between domains by aggregating the information from both word sequences and syntactic graphs. 
Specifically, GAST mainly contains two modules to learn comprehensive semantics. The first one is \emph{POS-based Transformer (POS-Transformer)}, which includes a new multi-head attention mechanism to encode the word semantics with the help of POS tags. The other is \emph{Hybrid Graph Attention (HGAT)}, which aims to learn and weigh the semantic influence between words and their neighbors with the help of the domain-invariant dependency relations. Finally, we propose a novel \emph{Integrated aDaptive Strategy (IDS)} which integrates an adversarial training and pseudo-label based semi-supervised learning to distill transferable semantic features. Extensive experiments on real-world datasets demonstrate the effectiveness of our proposed approach. 
In summary, the contributions of this work can be summarized as:
\begin{itemize}
    \item To the best of our knowledge, we present the first solution to address the CDSC problem by incorporating domain-invariant semantic knowledge from word sequences and syntactic graph structures simultaneously.
    \vspace{0.05cm}
  \item We propose a novel \emph{{G}raph {A}daptive {S}emantic {T}ransfer ({GAST})} model for syntactic graph learning. GAST contains a \emph{POS-Transformer} to learn sequential semantic from word sequences and POS tags, and a \emph{HGAT} to fully exploit adaptive syntactic knowledge with the help of dependency relations.
  \vspace{0.05cm}
  \item We further design an integrated adaptive strategy to optimize the transferability of the GAST model. Experimental results show that the proposed model achieves better results compared to other strong competitors.
\end{itemize}

%%%%%%%%%%%%%%%%%%  related work  %%%%%%%%%%%%%%%%%%%%
%%%%%%%%%%%%%%%%%%  related work  %%%%%%%%%%%%%%%%%%%%
%%%%%%%%%%%%%%%%%%  related work  %%%%%%%%%%%%%%%%%%%%
\section{related work} 
In the following, we will introduce two research topics which are highly relevant to this work, i.e., cross-domain sentiment classification and syntax modeling in NLP tasks.
\vspace{-0.2cm}
\subsection{Cross-domain Sentiment Classification}
Cross-domain sentiment classification (CDSC) aims to generalize a classifier that is trained on a source domain into a target domain in which labeled data is scarce. In the field of CDSC, a group of methods focuses on exploiting the explicit domain-shared knowledge~\cite{xia2011pos,mahalakshmi2015cross,pan2010cross}. Among them,  Blitzer et al. \shortcite{blitzer2006domain,blitzer2007biographies} proposed and extended the structural correspondence learning to identify the domain-shared features from different domains. Xia and Zong \shortcite{xia2011pos} designed an ensemble model to learn common features by using the part-of-speech tags. These earlier methods need to select domain-shared features manually while obtaining rich artificial features is a time-consuming and expensive process.

Recent years, many researchers studied the problem~\cite{glorot2011domain,chen2012marginalized,yu-jiang-2016-learning,ganin2016domain,li2018hierarchical,yuan2021learning} through the deep neural networks. Among them, Glorot et al.~\shortcite{glorot2011domain} first proposed a deep learning model named Stacking Denoising Autoencoder (SDA), which aimed to improve the scalability of high-dimensional data. Later, Chen et al.~\shortcite{chen2012marginalized} extended it as marginalized Stacked Denoising Autoencoder (mSDA). Along this line, 
Yu and Jiang \shortcite{yu-jiang-2016-learning} leveraged two auxiliary tasks to learn in-depth features together with a CNN-based classifier. 
Ganin et al. \shortcite{ganin2016domain} introduced a general domain adaptation strategy of the task by applying a gradient reversal layer. Ghosal et al. \cite{ghosal2020kingdom} enriched the semantics of a document by exploring the role of external commonsense knowledge. Li et al. \shortcite{li2017end,li2018hierarchical} incorporated the adversarial memory network and hierarchical attention transfer network into domain-adversarial learning to automatically identify invariant features. Zhang et al. \shortcite{zhang2019interactive} designed an interactive transfer network, which aims to extract interactive relations between sentences and aspects. Du et al. \shortcite{du2020adversarial} proposed a BERT-DAAT model and a post-training procedure to enforce the model to be domain-aware. Despite the promising results, most of these methods process sentences as whole word sequences and ignore the domain-invariant syntactic structures of the sentence.

Although some earlier studies have studied the syntactic information in CDSC~\cite{xia2011pos,mahalakshmi2015cross}, they only focused on modeling POS tags while largely ignoring the graph adaptive semantics behind the dependency relations. Thus, in this paper, we design a graph adaptive semantic transfer model to learn comprehensive semantics from both word sequences and syntactic structures. 

\vspace{-0.2cm}
\subsection{Syntax Modeling in NLP Tasks}
Syntactic information has been verified to be essential for many NLP tasks, such as aspect-based sentiment analysis (ABSA)~\cite{sun2019aspect,huang2019syntax,shi2019deep,zhang2021eatn,wang2020relational}. Among those methods, Huang et al.~\shortcite{huang2019syntax} utilized the syntactic information to represent the sentence as a graph structure instead of a word sequence. Wang et al.~\shortcite{wang2020relational} reshaped the dependency tree to learn aspect-aware semantics from syntactic information. These methods have indicated that syntactic information positively affects semantic representation. However, they are designed explicitly for aspect-based sentiment or other tasks, which can not perfectly apply to the CDSC task. For example, although numerous methods use syntactic information to help sentiment prediction in ABSA, this syntactic information supports the model to understand semantics rather than transfer semantics better. Thus, the performance of those approaches drops a lot under the domain adaptation scenario.

Furthermore, there are also some syntax-based studies in cross-domain aspect and opinion co-extraction task~\cite{li2012cross,ding2017recurrent,wang2018recursive}. However, most of those methods utilized the dependency relationships to generate an auxiliary label~\cite{ding2017recurrent} of the sentence or generate an auxiliary task for relation classification~\cite{wang2018recursive}, which is not in line with the purpose of CDSC. Though these models are not suitable for CDSC, they effectively prove the effectiveness of syntactic information in cross-domain tasks. Therefore, in this paper, we leverage syntactic structures as transferable information and incorporate it into the domain adaptation framework to learn domain-invariant features.

\begin{figure*} [t]
	\centering
	\includegraphics[width=1.88\columnwidth]{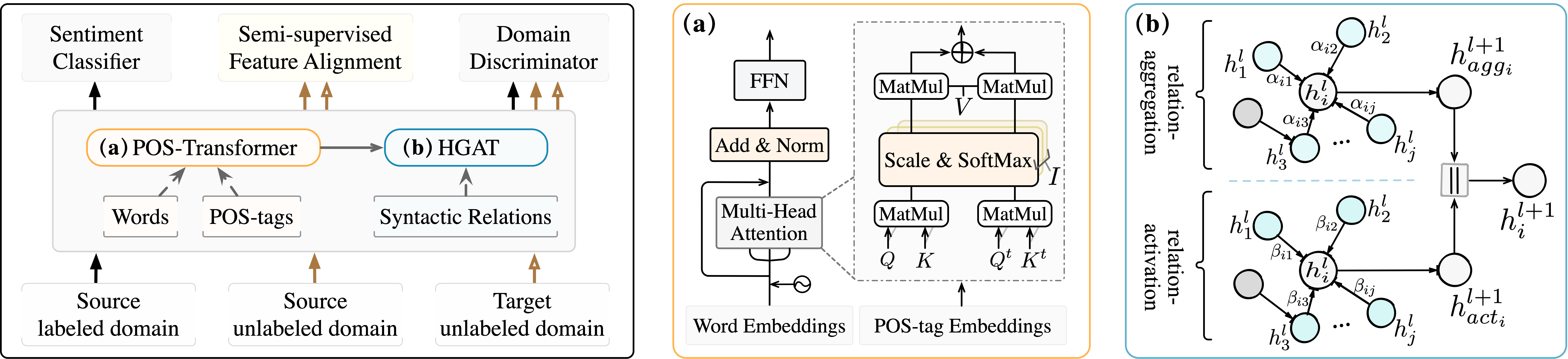}
	\caption{The architecture of {\textbf{GAST}}, which consists three parts: (a) the \emph{\textbf{POS-Transformer}} that can learn sequential semantic representation by considering both the word sequences and POS tags; (b) the \emph{\textbf{HGAT}} module which can exploit adaptive syntactic semantics of the sentence through the syntactic relation graph. (c) an \emph{\textbf{IDS}} (i.e., Sentiment Classifier, Semi-supervised Feature Alignment and Domain Discriminator) to optimize the model and encourage it to be domain-invariant and syntax-aware.
	}
	\label{fig:model}
\end{figure*}

However, though remarkable advance has been gained in the task, the above methods can only handle the sequential semantics in word sequences instead of the rich syntactic structure knowledge the sentence contains, which may not be able to learn transferable syntactic knowledge that is important to human. Unlike previous methods, our GAST can explore transferable semantics from syntactic structures and sequential information, thus can better simulate human's syntax rules. 

%%%%%%%%%%%%%% model %%%%%%%%%%%%%%%%
%%%%%%%%%%%%%% model %%%%%%%%%%%%%%%%
%%%%%%%%%%%%%% model %%%%%%%%%%%%%%%%
\section{METHODOLOGY}
%We now present the proposed GAST in detail. First, we introduce the problem definition. Then, we describe each component of GAST along with the advantages of such a design.

\subsection {Problem Definition}
Suppose we have two domains,  ${{\mathcal{D}}}_s$ is the source domain (i.e., contains labeled data $\{x_s^i, y_s^i\}_{i=1}^{n_{sl}} \in {{\mathcal{D}}}_s^l$ and unlabeled data $\{x_s^i\}_{i=n_{sl}+1}^{n_s} \in {\mathcal{D}}_s^u$) and ${\mathcal{D}}_t=\{x_t^i\}_{i=1}^{n_t}$ is the unlabeled target domain. $x_*^*$, $y_*^*$ denote samples and the corresponding label. The notations $n_{sl}$, $n_s$ and $n_t$ are the number of labeled data in source domain, the number of data in source domain and the number of data in target domain, respectively. Our goal is to learn a robust classifier from samples in the source and target domain and adapt it to predict the sentiment polarity of unlabeled examples in the target domain.

\subsection {Syntactic Graph and Embedding} 
\label{graph}
In the task, each input sentence $s$ contains \emph{n} words marked as $s={\{s_1, s_2, ..., s_n\}}$. To learn a syntax-aware representation, we transform each sentence into a syntactic dependency tree \emph{T} using an off-the-shelf dependency parser\footnote{https://github.com/allenai/allennlp.}~\cite{dozat2016deep}. Note that, the dependency tree can be represented as a syntax graph $\mathcal{G} = (\mathcal{V}, \mathcal{A}, \mathcal{R})$, where $\mathcal{V}$ includes all words of the sentence, $\mathcal{A}$ is adjacent matrix with $A_{ij}$ = 1 if there exists a dependency relation between word $s_i$ and $s_j$, and $A_{ij}$ = 0 otherwise. $\mathcal{R}$ is a set of syntactic relations (e.g., \emph{det}, \emph{nsubj} and \emph{cop}), where $R_{ij}$ corresponds to the relation label of $A_{ij}$. 

For our model, we conduct two types of word embedding methods: the {GloVe}~\cite{pennington2014glove} and pre-trained BERT embedding. Specifically, for GloVe embeddings, we map each word over a GloVe word matrix to get the vector. 
For BERT embeddings, we use an off-the-shelf toolkit\footnote{https://bert-as-service.readthedocs.io/en/latest/.} to encode each word and obtain the embeddings for each word instead of GloVe. For simplicity, we leverage $v_i \in \mathbb{R}^d$ as the representation vector of the $i$-th word in sentence, and the original word sequence ${s}$ is transformed into an embedding matrix, i.e., $E \in \mathbb{R}^{n \times d}$. Moreover, we encode each word's POS tag to an embedding vector $t_i \in \mathbb{R}^{d_t}$ and transform each syntactic relation label $R_{ij}$ to a vector $r_{ij} \in \mathbb{R}^{d_r}$, where $d$, $d_t$ and $d_r$ are the dimension of different embedding spaces.

\subsection{POS-Transformer} 
Since the superiority of the Transformer model in various sequential tasks~\cite{vaswani2017attention}, we utilize transformer-based encoder to learn semantic knowledge from the word sequence as well as the POS tags. As shown in Figure~\ref{fig:model} (a), the basic of our POS-Transformer is a new multi-head attention that creatively incorporates POS tags along with the traditional word sequences.

To be specific, for each attention head $i \in [1, I]$, we project the word's embedding matrix $E$ into the query, key, and value matrices, denoted as ${Q_i}$, ${K_i}$, ${V_i}$. Apart from word's embeddings, we also map the whole tag embedding matrix $E^t$ into matrices $Q_i^t$ and $K_i^t$, while keeping the value $V_i$ in this part to explore the interactive influence between POS tags and the words.
Then, we incorporate external POS tags knowledge along with word's sequential information to learn POS-based semantic representation (i.e., $Z$) of the sentence:
\begin{equation}
	\begin{aligned} 
	Z = concat(z_{1}, z_2, \ldots, z_I),
	 \end{aligned}
\end{equation}
\begin{equation}
\label{two}
	\begin{aligned} 
	z_i = Att. (Q_i, K_i, V_i) + Att. (Q_i^t, K_i^t, V_i),
	  \end{aligned}
\end{equation}
\begin{equation}
	\begin{aligned} 
	Att. (Q, K, V) = \operatorname{softmax}\left(\frac {Q K^{T}}{\sqrt{d / I}}\right) V,
	  \end{aligned}
\end{equation}
where $I$ denotes the number of attention head in transformer, $Att.( )$ is the attention function. %, which can be described as:
After getting the deep latent representation $Z$, we apply non-linear transformations on it and get the final output feature $R$ of the POS-Transformer module:\begin{equation}
	\begin{aligned}
	 R = \max(0, Z {W}_{1}+{b}_{1}) {W}_{2}+{b}_{2},
	\end{aligned}
\end{equation}
where ${W}_{1}$, ${W}_{2}$, ${b}_{1}$, ${b}_{2}$ are the weight and bias parameters. Through the above procedure, the POS-Transformer can calculate the correlation between each word with the help of POS tags. Therefore, the sequential semantic knowledge hidden in the word sequences and POS tags can be fully extracted.   

%\begin{figure}% [h]
%\vspace{-0.38cm}
%	\centering
%	\includegraphics[width=0.7\columnwidth]{figure/pos-transformer.pdf}
%	\caption{The architecture of {\textbf{POS-Transformer}}.
%	}
%	\label{fig:pos}
%	\vspace{-0.2cm}
%\end{figure}

%Relation-aware 
\subsection{Hybrid Graph Attention} 
In order to effectively learn syntactic graph embedding with full consideration of the syntactic dependencies, we adopt GAT to learn the relational features of the sentence. Generally, given a word $w_i$ (i.e., node $i$) with its deep hidden representation $h_i^l$ at $l$-th layer. GAT updates the node's hidden state (i.e., $h_i^{l+1}$) at $l$+1 layer by calculating a weighted sum of its neighbor states through the masked attention (i.e., compute $w_j$ for nodes $j \in \mathcal{N}_i$, where $\mathcal{N}_i$ is the neighborhood of node $i$ in the syntactic graph). 

However, the vanilla GAT uses an adjacent matrix as structural information, thus omitting dependency relations. Unlike vanilla GAT, we design a new Hybrid GAT to enhance information exchange among words via syntactic relations. As Figure~\ref{fig:model} (b) shows, HGAT contains two different calculation methods for better relation representation (i.e., relation-aggregation and relation-activation). The first is the relation-aggregate function, which is designed to learn transferable syntactic relations during the aggregate process. Thus, we could conduct the following formulas:
\begin{equation}
	h_{agg_i}^{l+1}=\|_{k=1}^{\bar{K}}\ \sigma (\sum_{j \in \mathcal{N}_{i}} \alpha_{i j}^{l k} {{W}}_{l k} h_{j}^{l}),
\end{equation}
\begin{equation}
\label{att1}
	f_{i j}^{l k} = \sigma(a_{l k}^{T}[{W}_{l k} h_{i}^{l} \| {W}_{l k} h_{j}^{l} \| {W}_{l k} r_{ij}]),
%f\left(a_{l k}^{T}\left[{W}_{l k} h_{i}^{l} \| {W}_{l k} h_{j}^{l} \| {W}_{l k} r_{ij}\right]\right)
\end{equation}
\begin{equation}
	\alpha_{i j}^{l k}=\frac{\exp \left(f_{i j}^{l k}\right)}{\sum_{j=1}^{\mathcal{N}_{i}} \exp \left(f_{i j}^{l k}\right)},
\end{equation}
where $\|$ denotes the concatenation of vectors, ${W}_{l k}$ is a learnable transformation matrix and $\bar{K}$ is the number of attention head in GAT. Besides, $a_{l k}^{T}$ is a learnable parameter and $\alpha_{i j}^{l k}$ is the attention coefficient in the $k$-th head at $l$-th layer. $\sigma( )$ is the \emph{LeakyReLU} function. $r_{ij}$ represents the syntactic relation embedding between word $i$ and word $j$.

%\begin{figure}% [h]
%\vspace{-0.38cm}
%	\centering
%	\includegraphics[width=0.81\columnwidth]{figure/hgat.pdf}
%	\caption{The architecture of {\textbf{HGAT}}.
%	}
%	\label{fig:hgat}
%	\vspace{-0.2cm}
%\end{figure}

As shown in Equation~\ref{att1}, the above relation-aggregate function can learn syntactic features by splicing the representation of both syntactic relation and nodes. However, the splicing operation is relatively intuitive and straightforward, which may not be able to capture the interactive influence between nodes and their syntactic dependency relations. To this end, we calculate the activation probability of each syntactic dependency relation by leveraging the scaled dot-product attention~\cite{vaswani2017attention} so that the adaptive impact of different syntactic relations can be  reassigned and further explored. Specifically, in our implementation, the relation-activation function is represented as:
\begin{equation}
\label{eight}
	\beta_{i j}^{l k}=\frac{\exp \left(F_{act.} (h_i^l, h_j^l)\right)}{\sum_{j=1}^{\mathcal{N}_{i}} \exp \left(F_{act.} (h_i^l, h_j^l)\right)},
\end{equation}
\begin{equation}
	F_{act.} = \frac{\left(W_{Q}^{lk} h_{i}^{l}\right)\left(W_{K}^{lk} h_{j}^{l}+W_{Kr}^{l} r_{i j}\right)^{T}}{\sqrt{d / \bar{K}}},
\end{equation}
\begin{equation}
	h_{act_i}^{l+1}=\|_{k=1}^{\bar{K}}\ \sigma (\sum_{j \in \mathcal{N}_{i}} \beta_{i j}^{l k} ({{W}}_V^{l k} h_{j}^{l} + {{W}}_{Vr}^{l} r_{ij})),
\end{equation}
where $W_{Q}^{lk}$, $W_{K}^{lk}$, $W_{V}^{lk}$,  $W_{Kr}^{l}$ and $W_{Vr}^{l}$ are learnable parameter matrices which are shared across attention heads. With the relation-attentional function, the weight $\beta_{i j}^{l k}$ greatly enhances the explanatory ability of the model and enables  GAST to learn crucial features from the neighbors with high relation scores in the syntactic graph.

Through the above two relational  functions, we can obtain two syntax-enhanced word representations. In order to better improve the comprehensiveness of syntactic information representation, we generate the final representation of each word at $l$+1 layer through: 
\begin{equation}
	h_{i}^{l+1} = h_{agg_i}^{l+1}\ \| \ h_{act_i}^{l+1}.
\end{equation}

The sentence's final representation $H$ is the average pooling result of each word representation in the sentence. Since HGAT is able to calculate the correlation between neighbor words in the adaptive syntactic graph, the transferable semantic features contained in the syntactic structures can be fully learned. 

\subsection{Integrated Adaptive Strategy} 
\label{IAS}
In this subsection, we introduce how GAST obtains the syntax-aware and domain-invariant features through an integrated adaptive strategy, which mainly includes three loss components. %///
As shown in Figure~\ref{fig:model}, the strategy includes three loss functions: a classifier loss for sentiment knowledge learning, a discriminator loss for invariant feature extracting across domains, as well as a syntax feature alignment loss for syntax-aware feature alignment.

\vspace{0.05cm}
\textbf{Sentiment Classifier}.
The sentiment classifier is simply defined as $\hat{y}_s=softmax({{W}_s H + {b}_s})$, which is used to classify sentiment polarities. The objective loss function of the classifier is defined as:
\begin{equation}
%\small
	L_{c}=-\frac{1}{n_{s}^{l}} \sum_{i=1}^{n_{s}^{l}}({y}_{s}^i \ln \hat{y}_{s}^i+(1-{y}_{s}^i) \ln (1-\hat{y}_{s}^i)),
\end{equation}
where ${y}_s^i$ is the ground-truth for the $i$-th sample in the source labeled domain ${\mathcal{D}}_s^l$. ${W}_s$ and ${b}_s$ represent learnable parameters. The classifier loss $L_c$ will train the model to learn better sentiment-aware features from both sequential information and syntactic structures.

\vspace{0.05cm}
\textbf{Domain Discriminator}.
Cross-domain sentiment classifiers perform well when the learned features are domain-invariant. To this end, we devise a domain discriminator, which aims to facilitate knowledge transfer between domains. Specifically, we feed the feature $H$ into the {softmax} layer for domain classification $\hat{y}_d=softmax({{W}_d H + {b}_d})$. The optimization goal is to train a model that can fool the discriminator so that the learned features can be transferred from domain-specific to domain-invariant~\cite{lample2018unsupervised,sun2019aspect}. Therefore, we reverse the domain label in the training process and the optimization objective function can be defined as:
\begin{equation}
%\small
	L_{d}=-\frac{1}{N} \sum_{i=1}^{N}({y}_{d}^i \ln \hat{y}_{d}^i+(1-{y}_{d}^i) \ln (1-\hat{y}_{d}^i)),
\end{equation}
where $y_d^i$ is the reversed ground-truth of sample $i$ (i.e., swap the sample's domain label). $N$ is the sum of $n_s$  from the source domain and $n_t$ is comes from the target domain.

\vspace{0.05cm}
\textbf{Semi-supervised Learning (SSL)}.
Since there are a large amount of unlabeled data in the source and target domain, it is hard to train an optimal classifier. Fortunately, syntactic information plays a critical role in semantic representation which is helpful for sentiment classification~\cite{huang2019syntax,wang2020relational}. Thus, we utilize the syntactic information of unlabeled data in both domains to estimate the sentiment label so that GAST can eliminate the sentiment discrepancies between domains (i.e., explore and align sentiment features via syntactic-semantic information) via semi-supervised learning~\cite{he2018adaptive}. 

Specifically, we first attempt to estimate the sentiment “pseudo” label through semi-supervised learning. Inspired by~\cite{li2020cross}, we minimize the entropy penalty to disambiguate the positive and negative samples over the unlabeled data from both domains. The loss function is:
\begin{equation}
	L_{a}=-\frac{1}{M} \sum_{i=1}^{M} \sum_{j=1}^{C} \tilde{y}^i \ln \tilde{y}^i \ ,
\end{equation}
where $\tilde{y}^i$ is the label distribution estimated by our model, $C$ is the number of categories and $M$ is the sum of $n_s-n_{sl}$ and $n_t$. Note that, the domain discrepancy can be effectively reduced through feature alignment~\cite{he2018adaptive,li2020cross}.

\subsection{Model Training}
Since there are three objective functions in the model, we conduct an integrated strategy to jointly optimize the final loss. The formulation is defined as:
\begin{equation}
	\begin{split}
	&L\ =\ {\lambda}_{c} L_{c}\ +\ {\lambda}_{d} L_{d}\ +\ {\lambda}_{a} L_{a}\ ,
	\end{split}
	\label{eq:loss_all}
\end{equation}
where $\lambda_{c}$, $\lambda_{d}$ and $\lambda_{a}$ are hyper-parameters to balance different objective losses. The training goal is to minimize the integrated loss $L$ with respect to the model parameters. Additionally, all the parameters are optimized by the standard back-propagation algorithm.

%%%www
%where $\lambda_{c}$, $\lambda_{d}$ and $\lambda_{a}$ are hyper-parameters to balance different objective losses. The training goal is to minimize the integrated loss $L$ with respect to the model parameters. Additionally, all the parameters are optimized by the standard back-propagation.

\subsection{Summary and Remarks}
It should be noted that in this paper we focus on learning domain-invariant semantics from both sequential texts and adaptive syntactic structures simultaneously. We optimize the model with an integrated adaptive strategy, which deeply explores the impacts of the adaptive graph structures for cross-domain sentiment classification. Although there have been some studies (e.g., in the ABSA task~\cite{wang2020relational,gong2020unified,meijer2021explaining,zhang2021eatn}) to utilize syntactic graph structures to enhance semantic representation, the transferability exploration for adaptive graph structure information is still limited. That is, our proposals are the first solution to learn domain adaptive graph semantics for CDSC, and we encourage more effective studies to be explored and further improve our graph adaptive framework.

\section{Experiments}
\subsection{Dataset Setup}%~\cite{blitzer2007biographies}
We evaluate GAST on four widely-used Amazon datasets, i.e., DVD ($\bm{D}$), Book ($\bm{B}$), Electronics ($\bm{E}$) and Kitchen ($\bm{K}$). As shown in Table~\ref{tab:data}, for each domain, there are 2,000 labeled reviews (i.e., 1,000 positive and 1,000 negative) as well as 4,000 unlabeled reviews. We follow the dataset configurations as previous studies~\cite{du2020adversarial,li2017end,li2018hierarchical}, that is, we randomly choose 800 positive and 800 negative reviews from the source domain as training data, the rest 400 as validation data to find the best hyperparameters, and all labeled reviews from the target domain for testing.

\begin{table} 
    \vspace{0.06cm}
    \caption{Statistics of datasets after pre-processing.} 
    \label{tab:data} 
    \centering
    \begin{tabular}{p{1.88cm}<{\centering}|p{1.2cm}<{\centering}|p{1cm}<{\centering}|p{1.2cm}<{\centering}|p{1.2cm}<{\centering}}
            \toprule
            \multirow{2}{*}{Domains} &\multicolumn{3}{c}{Testing set percentage}\\ 
            \specialrule{0em}{1pt}{0pt}
            \cline{2-5}\specialrule{0em}{2pt}{0pt}& \#Train&\#Vali.&\#Test&\#Unlabel\\
            \midrule
            Books &1,600&400 & 2,000&4,000\\
            DVD &1,600&400 & 2,000&4,000 \\
            Electronics & 1,600&400 &2,000&4,000\\
            Kitchen & 1,600&400 &2,000& 4,000\\
            \bottomrule
    \end{tabular}
\end{table}

\begin{figure}
	\centering
	\vspace{0.05cm}
	\includegraphics[width=0.96\columnwidth]{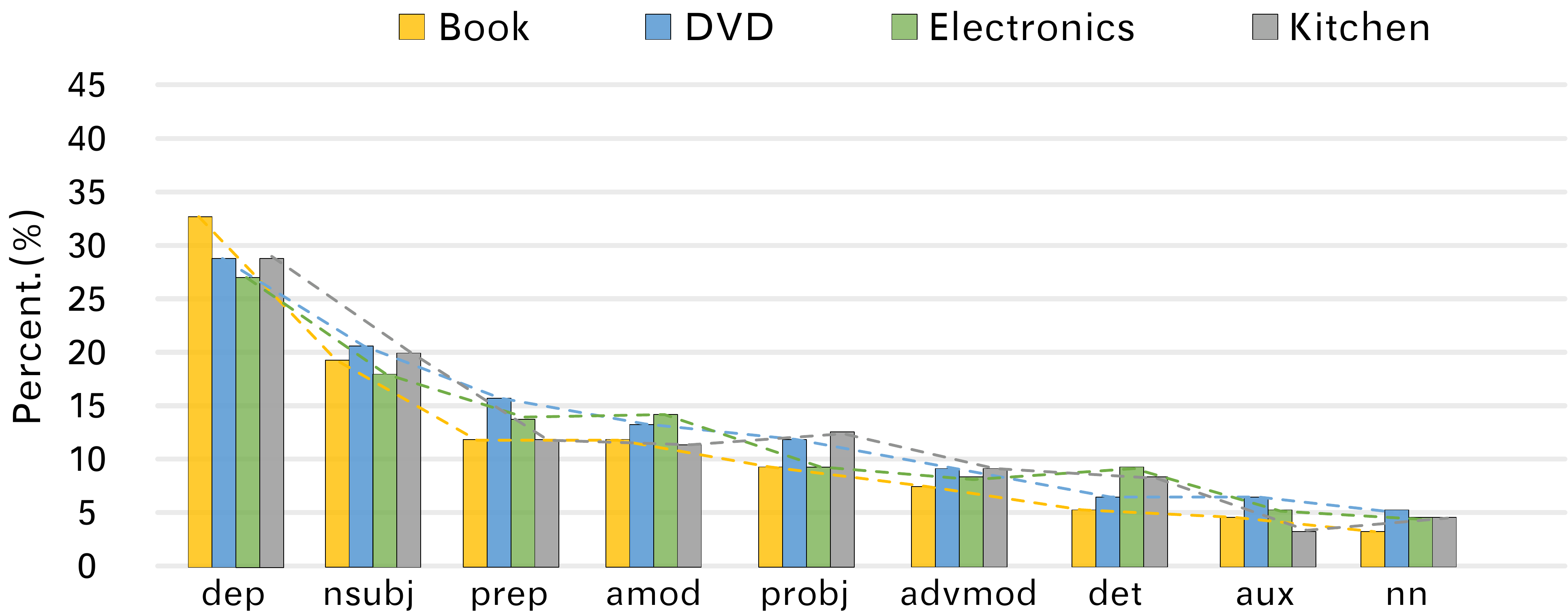}
	\vspace{-0.1cm}
	\caption{The percent of transferable dependency relations in different domains. We visualized the top 9 relations.
	}
	\label{fig:data}
	\vspace{-0.1cm}
\end{figure}

\subsection{Data Analysis}
In this subsection, we count the ratio of different syntactic relationships as illustrated in Figure~\ref{fig:data}. We can observe some phenomena intuitively. First, for each syntactic dependency relation, the proportions between various domains are close, meaning each sentence’s components might be remarkably similar, even in different domains. Besides, the curve (i.e., dotted lines) of the syntactic relations is identical for each domain. The statistical observations above may indicate that the syntactic structures of the sentence are domain-invariant between domains from the perspective of data mining.

\subsection{Baseline Methods}
We compare the GAST model with multiple representative transfer baselines as well as some non-transfer approaches, which have achieved significant performance in recent years. The methods are listed below.

\begin{itemize}
	\item \textbf{SCL}~\cite{blitzer2006domain} is a linear method, which aims to solve feature the mismatch problem by aligning domain common and domain unique features.
	\item \textbf{SFA}~\cite{pan2010cross} is a method which aims to build a bridge between the source and the target domains by aligning common and unique features.
	\item \textbf{mSDA}~\cite{chen2012marginalized} is proposed to automatically learn a unified feature representation for sentences from a large amount of data in all the domains.
	\item \textbf{DANN}~\cite{chen2012marginalized} is based on adversarial training. DANN performs domain adaptation with the representation encoded in a 5000-dimension feature vector.
	\item \textbf{AMN}~\cite{li2017end} is a method which learns domain-shared features based on memory network sentiment.
	\item \textbf{HATN}~\cite{li2018hierarchical} is a hierarchical attention transfer network which is designed to focus on both of the word-level and sentence-level sentiment. 
	\item \textbf{IATN}~\cite{zhang2019interactive} is an interactive attention transfer model which focus on mining the deep interactions between the context words and the aspects.
	\item \textbf{BERT-DAAT}~\cite{du2020adversarial} contains a post-training and an adversarial training process which aims to inject target domain knowledge to BERT and encourage it to be domain-aware.
\end{itemize}

Besides, we also borrow three competitive non-transfer deep representation learning method for comparison, i.e., Naive {{LSTM}}~\cite{hochreiter1997long}, {{TextGCN}}~\cite{yao2019graph} and {{FastGCN}}~\cite{chen2018fastgcn}. The detail is illustrated as follow.

\begin{itemize}
	\item \textbf{LSTM}~\cite{hochreiter1997long} utilizes neural network to learn the hidden states and obtain the averaged vector through mean pooling to predict the sentiment polarity.
	\item \textbf{TextGCN}~\cite{yao2019graph} a simple and effective graph neural network for text classification that captures high-order neighborhoods information from the syntactic graph.	
	\item \textbf{FastGCN}~\cite{chen2018fastgcn} a fast improvement of the GCN model for learning graph embeddings. Here, we perform it on our syntactic graph to learn syntax-aware sentiment features.
\end{itemize}

\begin{table*} 
\linespread{0.86} 
\small
    \caption{Sentiment classification accuracy (\%) on the twelve transfer tasks. } 
    \label{tab:result} 
    \centering
    \resizebox{\textwidth}{!}{
        \begin{tabular}{p{2cm}|p{0.66cm}<{\centering}p{0.66cm}<{\centering}p{0.66cm}<{\centering}|p{0.66cm}<{\centering}p{0.66cm}<{\centering}p{0.66cm}<{\centering}|p{0.66cm}<{\centering}p{0.66cm}<{\centering}p{0.66cm}<{\centering}|p{0.66cm}<{\centering}p{0.66cm}<{\centering}p{0.66cm}<{\centering}}
            \toprule
            \specialrule{0em}{2pt}{0pt}
            \multirow{2.8}{*}{\textbf{Baselines}} 
            &\multicolumn{3}{c}{\textbf{\emph{DVD} ($\bm{D}$)}} &\multicolumn{3}{c}{\textbf{\emph{Book} ($\bm{B}$)}} &\multicolumn{3}{c}{\textbf{\emph{Electronics} ($\bm{E}$)}} &\multicolumn{3}{c}{\textbf{\emph{Kitchen} ($\bm{K}$)}}\\ 
            \specialrule{0em}{2pt}{0pt}
            \cline{2-4}
            \cline{5-7}
            \cline{8-10}
            \cline{11-13}
            \specialrule{0em}{3pt}{0pt}
            &$\bm{D}$$\mapsto $$\bm{B}$
            &$\bm{D}$$\mapsto $$\bm{E}$
            &$\bm{D}$$\mapsto $$\bm{K}$
            &$\bm{B}$$\mapsto $$\bm{D}$
            &$\bm{B}$$\mapsto $$\bm{E}$
            &$\bm{B}$$\mapsto $$\bm{K}$
            &$\bm{E}$$\mapsto $$\bm{D}$
            &$\bm{E}$$\mapsto $$\bm{B}$
            &$\bm{E}$$\mapsto $$\bm{K}$            
            &$\bm{K}$$\mapsto $$\bm{D}$
            &$\bm{K}$$\mapsto $$\bm{B}$
            &$\bm{K}$$\mapsto $$\bm{E}$           
            \\
            \specialrule{0em}{2pt}{0pt}
            \toprule
            \specialrule{0em}{2pt}{0pt}            
            SCL 
            &\ 77.8%$^{\dagger}$
            &\ 75.2%$^{\dagger}$
            &\ 75.5
            &\ 80.4%$^{\dagger}$
            &\ 76.5%$^{\dagger}$
            &\ 77.1
            &\ 74.5%$^{\dagger}$
            &\ 71.6%$^{\dagger}$
            &\ 81.7
            &\ 75.2
            &\ 71.3
            &\ 78.8     
            \\
                   
            SFA 
            &\ 78.8%$^{\dagger}$
            &\ 75.8%$^{\dagger}$
            &\ 75.7%$^{\ddagger}$
            &\ 81.3%$^{\dagger}$
            &\ 75.6%$^{\dagger}$
            &\ 76.9%$^{\ddagger}$
            &\ 75.4%$^{\dagger}$
            &\ 72.4%$^{\dagger}$
            &\ 82.6%$^{\ddagger}$
            &\ 74.7%$^{\ddagger}$
            &\ 72.4%$^{\ddagger}$
            &\ 80.7%$^{\ddagger}$
            \\
            
            DANN 
            &\ 80.5%$^{\dagger}$
            &\ 77.6%$^{\dagger}$
            &\ 78.8%$^{\ddagger}$
            &\ 83.2%$^{\dagger}$
            &\ 76.4%$^{\dagger}$
            &\ 77.2%$^{\ddagger}$
            &\ 77.6%$^{\dagger}$
            &\ 73.5%$^{\dagger}$
            &\ 84.2%$^{\ddagger}$
            &\ 75.1%$^{\ddagger}$
            &\ 74.3%$^{\ddagger}$
            &\ 82.2%$^{\ddagger}$
            \\ 
            
            AMN 
            &\ 84.5%$^{\dagger}$
            &\ 81.2%$^{\dagger}$
            &\ 82.7%
            &\ 85.6%$^{\dagger}$
            &\ 82.4%$^{\dagger}$
            &\ 81.7%$^{\ddagger}$
            &\ 81.7%$^{\dagger}$
            &\ 76.6%$^{\dagger}$
            &\ 85.7%$^{\ddagger}$
            &\ 81.5%
            &\ 80.9%$^{\ddagger}$
            &\ 86.1%$^{\ddagger}$
            \\
            
            HATN 
            &\ 86.6
            &\ 86.3
            &\ 87.4
            &\ 86.5
            &\ 85.7
            &\ 86.8
            &\ 84.3
            &\ 81.5
            &\ 87.9
            &\ 84.7
            &\ 84.1
            &\ 87.0
            \\
            
            IATN 
            &\ 87.0
            &\ 86.9
            &\ 85.8
            &\ 86.8
            &\ 86.5
            &\ 85.9
            &\ 84.1
            &\ 81.8
            &\ 88.7
            &\ 84.4
            &\ 84.7
            &\ 87.6
            \\
            BERT-DAAT
            &\ 90.8
            &\ 89.3
            &\ 90.5
            &\ 89.7
            &\ 89.5
            &\ 90.7
            &\ 90.1
            &\ 88.9
            &\ 93.1
            &\ 88.8
            &\ 87.9
            &\ 91.7
            \\
            \specialrule{0em}{1pt}{0pt}
            \midrule            
            \specialrule{0em}{2pt}{0pt}
             LSTM
            &\ 75.6%$^{\dagger}$
            &\ 73.4%$^{\dagger}$
            &\ \ \ -
            &\ 78.6%$^{\dagger}$
            &\ 75.2%$^{\dagger}$
            &\ \ \ -
            &\ 72.2%$^{\dagger}$
            &\ 69.6%$^{\dagger}$
            &\ \ \ -
            &\ \ \ -
            &\ \ \ -
            &\ \ \ -
            \\
            TextGCN 
            &\ 80.8
            &\ 77.6
            &\ 79.2
            &\ 85.3
            &\ 81.1
            &\ 79.7
            &\ 82.6
            &\ 78.2
            &\ 82.3
            &\ 83.3
            &\ 84.1
            &\ 81.7
            \\
            FastGCN 
            &\ 81.6
            &\ 80.6
            &\ 81.1
            &\ 86.0
            &\ 82.7
            &\ 82.0
            &\ 83.5
            &\ 78.7
            &\ 84.5
            &\ 84.2
            &\ 85.7
            &\ 83.4
            \\
            \specialrule{0em}{1pt}{0pt}
            \midrule
            \specialrule{0em}{2pt}{0pt}
             {GAST} 
             &\ {87.9}
             &\ {87.3}
             &\ {89.1}
             &\ {88.2}
             &\ {86.2}
             &\ {87.4}
             &\ {85.6}
             &\ {83.4}
             &\ {89.3}
             &\ {87.7}
             &\ {87.5}
             &\ {89.4}
             \\
             \textbf{BERT-GAST} 
             &\ \textbf{91.1}
             &\ \textbf{90.7}
             &\ \textbf{92.1}
             &\ \textbf{90.4}
             &\ \textbf{91.2}
             &\ \textbf{91.5}
             &\ \textbf{90.7}
             &\ \textbf{89.4}
             &\ \textbf{93.5}
             &\ \textbf{89.7}
             &\ \textbf{89.2}
             &\ \textbf{92.6}
            \\
            \specialrule{0em}{2pt}{0pt}
            \midrule 
            \midrule 
            \specialrule{0em}{2pt}{0pt}
            G\_\emph{Non\_Pos-Tran.}
            &\ 85.9
            &\ 84.7
            &\ 87.6
            &\ 86.8
            &\ 83.4
            &\ 85.5
            &\ 84.2
            &\ 80.4
            &\ 87.8
            &\ 85.8
            &\ 85.5
            &\ 87.4\\
            G\_\emph{Non\_HGAT} 
            &\ 86.6
            &\ 85.9
            &\ 88.1
            &\ 87.4
            &\ 85.0
            &\ 86.1
            &\ 84.5
            &\ 81.3
            &\ 88.2
            &\ 86.4
            &\ 86.7
            &\ 88.2
            \\
            G\_\emph{Non\_IDS} 
            &\ 87.2
            &\ 86.6
            &\ 87.9
            &\ 87.6
            &\ 85.8
            &\ 86.7
            &\ 85.0
            &\ 82.6
            &\ 88.5
            &\ 85.9
            &\ 86.2
            &\ 87.7
            \\  
            G\_\emph{Non\_agg} 
            &\ 87.5
            &\ 86.7
            &\ 88.9
            &\ 88.0
            &\ 85.9
            &\ 86.9
            &\ 85.2
            &\ 82.6
            &\ 89.0
            &\ 87.3
            &\ 87.2
            &\ 89.1
             \\
            G\_\emph{Non\_act} 
            &\ 87.3
            &\ 86.3
            &\ 88.7
            &\ 87.7
            &\ 85.3
            &\ 86.2
            &\ 84.8
            &\ 81.8
            &\ 88.7
            &\ 86.9
            &\ 87.1
            &\ 88.7
            \\          
            \bottomrule
        \end{tabular}
    } 
\end{table*}

\subsection{Implementation Details}
In the experiments, we initialize the dimension of the GloVe embeddings to 300 and utilize the BERT-base uncased model (layer=12, head=12, hidden=768)~\cite{kenton2019bert} in the off-the-shelf toolkit. The dimension of POS tags ($d_t$) and syntactic relations ($d_r$) is initialized to 30. The number of attention heads $I$ and $\bar{K}$ are set to 8 and 3. We adopt Adam~\cite{kingma2015adam} as optimizer with learning rate 10$^{-4}$ and dropout rate as 0.25 to train the model with the batch size of 32.  The final settings of $\lambda_{c}$, $\lambda_{d}$ and $\lambda_{a}$ are 1, 1 and 0.8, respectively. All of the methods are implemented by Python and are trained on a Linux server with two 2.20 GHz Intel Xeon E5-2650 CPUs and four Tesla V100 GPUs.
Finally, follow most previous studies~\cite{li2018hierarchical,zhang2019interactive,du2020adversarial}, we use the widely used metric (i.e., accuracy) for the model evaluation. We will release the code and dataset once the paper is accepted.

\subsection{Experimental Results}
In this section, we evaluate the performance of our model on public datasets along with detailed analysis of results in Table~\ref{tab:result}. The major results are summarized as follows:
\begin{itemize}
	\item In most tasks, neural-based transfer models (e.g., DANN and IATN) perform better compared with the SCL and SFA, which only manually selects common features (i.e., ``pivots''). The phenomenon demonstrates the power of deep methods, especially the models specially designed for CDSC. Meantime, the graph-based models outperform the sequential model LSTM and some transferable models (e.g., SFA and DANN) a lot, proving that the graphical syntactic structure is important for cross-domain semantic representation.
	\vspace{0.035cm}
	\item We also observe that the performance of GAST outperforms most baseline methods. Specifically, comparing with those neural-based transfer models (e.g., AMN, HATN, and IATN), our GAST model outperforms most of them. We conjecture one possible reason is that the performance of GAST can be significantly improved by fully exploring graphical syntactic structures of the sentences, i.e., POS-tags and syntactic dependency relations. Besides, the comparison with the graph-based models (i.e., TextGCN and FastGCN) further highlights the superiority of GAST. We believe the reason is that GAST is able to consider the sequential information and syntactic structures jointly; hence, the comprehensive semantic features could be better encoded and learned.  
	\vspace{0.035cm}
	\item As Table~\ref{tab:result} shows, the pre-trained language model (PLM, i.e., BERT-DAAT)  can outperform all the existing CDSC methods and non-transfer approaches by significant margins, which proves the semantic extraction capabilities of the large-scale pre-trained language models in this task. Nevertheless, after incorporating the BERT embeddings, our proposed model (i.e., BERT-GAST) gets further improvement and achieves a new SOTA. It means that the performance of GAST can be further improved based on the advanced PLMs in the future.
\end{itemize}

In summary, all the evidence above indicate that GAST outperforms other strong baselines on diverse transfer tasks. These observations, meanwhile, imply that the domain-invariant semantics learned by the proposed model are more effective and transferable for cross-domain sentiment classification.

\subsection{Ablation Study}
\label{ab}
In this subsection, we conduct multiple ablation studies to verify the effect of different modules. In what follows, we first describe the variants of GAST and then analyze how each of them affects the final performance: 
\begin{itemize}
	\item \textbf{G\_\emph{Non\_Pos-Transformer}}: it utilizes the vanilla transformer instead of the POS-Transformer to learn the sequential feature representations, i.e., to eliminate the impact of POS-tag.
	\item \textbf{G\_\emph{Non\_HGAT}}: a variant of the proposed model which removes the HGAT module directly, i.e., to eliminate the effect of the syntactic dependency relations.
	\item \textbf{G\_\emph{Non\_IDS}}: a variant of the GAST model, which replaces integrated adaptive strategy with the classical GRL strategy\footnote{GRL is a famous and widely used adversarial strategy in cross-domain sentiment classification task. For space saving, we detailed this strategy in the supplementary material (i.e., Appendix A).}, i.e., to verify the effectiveness of IDS.
	\vspace{0.02cm}
	\item \textbf{G\_\emph{Non\_agg}} and \textbf{G\_\emph{Non\_act}}: two different variants which remove the relation-aggregate function and relation-activation function respectively to verify the effect of them.
\end{itemize}

 \begin{figure*}
	\centering
	\includegraphics[width=2\columnwidth]{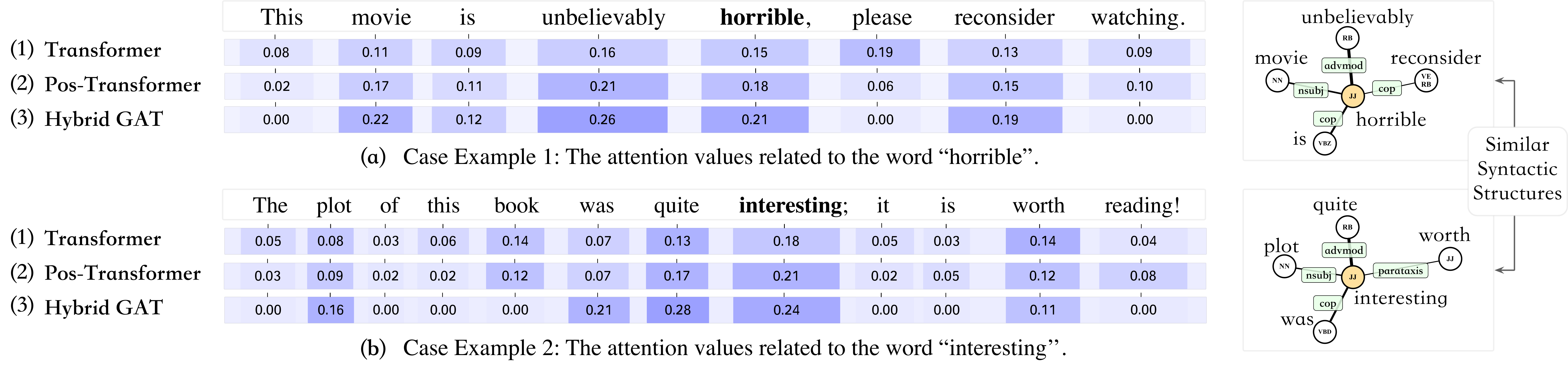}
	\vspace{-0.15cm}
	\caption{Attention score visualization of the different words. The attention values from vanilla attention (i.e., $Att.(Q, K, V$) in formula~\ref{two}), POS-attention (i.e., $Att.(Q^t, K^t, V$) in formula~\ref{two}) and HGAT (i.e., $\beta$ in formula~\ref{eight}) are associated with the row (1), row (2), and row (3) respectively in both examples. Note that, some values are infinitely close to 0. That makes sense because HGAT makes the attention value more concentrated on the syntactic-related words.
	}
	\label{fig:att}
\end{figure*}

The ablation results are shown in Table~\ref{tab:result}. To be specific, from the comparison results of G\_\emph{Non\_Pos-Transformer} and GAST, we can find that the performances decrease a lot when removing the POS-Transformer. It verifies the effectiveness of the POS-Transformer and demonstrates that the POS-tag information is significant. Besides, the results between G\_\emph{Non\_HGAT} and GAST indicate that HGAT can encode the transferable relational features effectually. Moreover, the performance of $G\_\emph{Non\_IDS}$ falls far behind the standard GAST, which validates that our proposed IDS is more effective than the traditional GRL strategy. Finally, after eliminating the impact of relation-aggregate (G\_\emph{Non\_agg}) and relation-activation (G\_\emph{Non\_act}) function, the performance drops in varying degrees, which further validates the importance of modeling the syntactic relations and the effectiveness of our two relational functions. In conclusion, the above ablation results demonstrate that our GAST is able to facilitate performance through multiple modules and gains superior prediction improvement in the CDSC task.

\subsection{Case Study}
\label{att}
To intuitively assess the effects of syntactic information, we visualize the attention scores in different layers of two examples from the DVD and Book domain. As Figure~\ref{fig:att} shows, the values in row (1) is attention score from the vanilla transformer, which means that the calculation of each word's attention does not consider POS tags and relational information. Comparatively, the attention values in row (2) and row (3) assess the POS tags and the syntactic relations, respectively. In the right part of Figure~\ref{fig:att}, we also show the syntactic structures of those two examples w.r.t the sentiment words.

As we can see from example 1 in Figure~\ref{fig:att}, the vanilla transformer makes extra decisions on some unrelated word (e.g., \emph{``this''}, \emph{``please''}) and pays much attention to these uncritical words. On the contrary, POS-transformer can alleviate this problem by revising attention scores with the help of POS tags. We believe one of the reasons is that the POS tags (e.g., ``JJ'' in the right part of Figure~\ref{fig:att}) can enforce the model to pay more attention to sentiment-related words no matter in which domain. Besides, HGAT could deal with the problem more appropriately through the domain-invariant syntactic relations between words, i.e., highlight the crucial neighbor words (e.g., \emph{``is''}, \emph{``unbelievably''}) via dependency relations as shown in right part of Figure~\ref{fig:att}. Finally, we can also observe similar phenomenon in example 2, which indicates that syntactic features are invariant and transferable between domains, as we mentioned before.

To sum up, the above case examples' visualization results convince us that the domain-invariant syntactic information is essential for cross-domain sentiment classification. Meantime, our proposed GAST model can capture essential graph adaptive semantics that is reasonably necessary for domain adaptation.

\subsection{Assessment of Adaptive Efficiency}
\begin{figure}
	\centering
	\includegraphics[width=0.98\columnwidth]{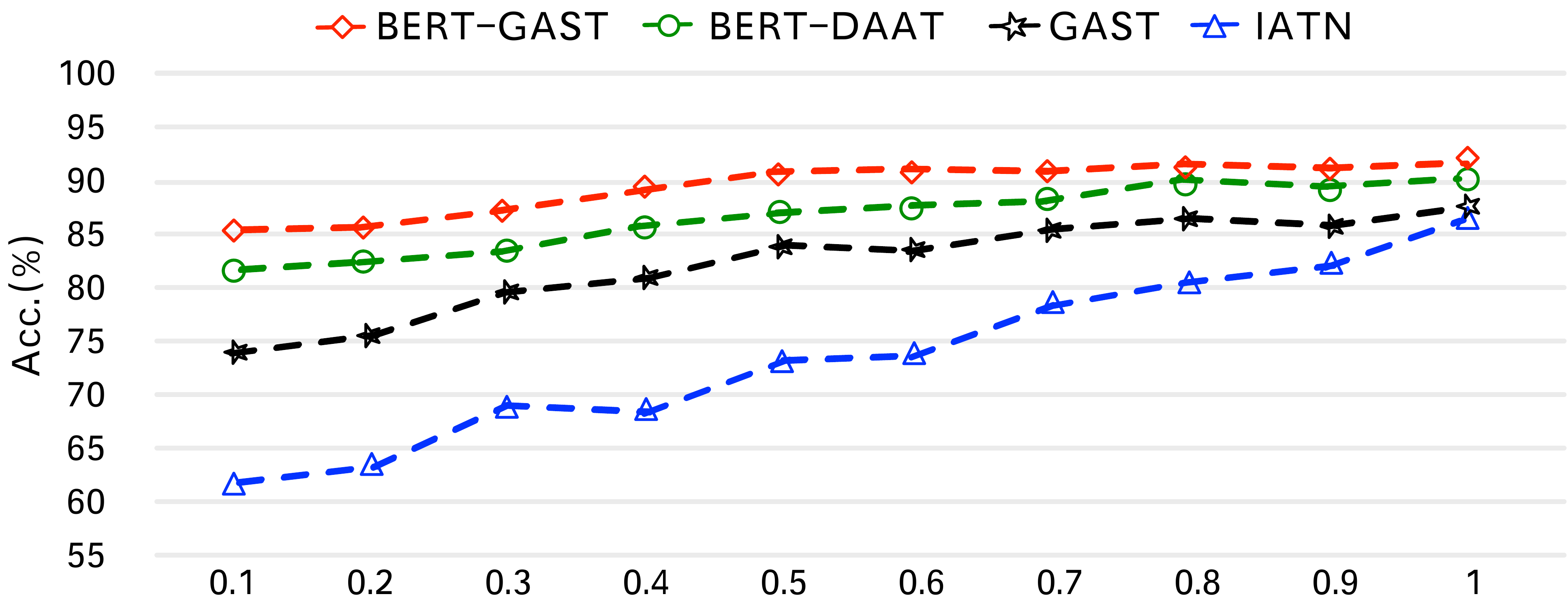}
	\vspace{-0.1cm}
	\caption{The influence of sample number. We explore the impact of sample number with different ratio (i.e., abscissa) of source domain. For the limited space, we only show the results of the task ``$\bm{B}$$\mapsto $$\bm{D}$''.
	}
	\label{fig:para}
	\vspace{-0.1cm}
\end{figure}

To study the adaptive efficiency of different models, we test several models' performance on the target domain with varying training sample rates. From the overall results in Figure~\ref{fig:para}, we can observe some interesting phenomena. First, we find that GAST can be trained with only 10\% samples, while IATN does not work well with such limited data. Meantime, GAST with 40\% samples even performs better than IATN with 80\% samples. All these observations denote that GAST gains an advanced adaptive ability and efficiency, thus significantly reducing the number of training samples. Second, BERT-GAST obtains superior results under all dataset scales, while the performance of other baseline models (i.e., BERT-DAAT, GAST, and IATN) is relatively low. This observation indicates that the representations learned from our BERT-GAST contain much more transferable sentiment knowledge than other baseline methods.

\subsection{Influence of Adaptive Syntactic Graph}
As we mentioned before, the syntactic structure plays a critical role in our proposed method. To evaluate the impact of different syntactic features, inspired by~\cite{wang2020relational}, we conduct a comparative experiment using two well-known dependency parsers, i.e., Stanford Parser~\cite{chen2014fast} and Biaffine Parser\footnote{https://github.com/allenai/allennlp.} ~\cite{dozat2016deep}, to construct our syntactic graph. The performance (i.e., accuracy, \%) of these two category graphs on $\bm{D}$$\mapsto $$\bm{*}$ tasks is shown
in Table~\ref{tab:parser}. The value in parentheses represents an absolute improvement, and method (1) indicates the {G\_\emph{Non\_HGAT}} which is described in section~\ref{ab}.
From the results, we can easily draw some conclusions. First, both two dependency parsers have made significant improvements compared with the method (1), the accuracy improvement on different transfer tasks is between 1.0\% and 1.4\%.
Second, the quality of the syntactic graph does affect the final performance of the model, which indicates the effectiveness of syntactic features for CDSC research.
Finally, it also implies that although existing parsers can capture most of the syntactic information correctly, our GAST's performance has potential to be further improved with the advances of parsing techniques in the future.

\begin{table}
\linespread{0.88} 
    \centering
        \caption{The performance (\%) of different syntactic graphs constructed by different parsers on $\bm{D}$$\mapsto $$\bm{*}$ tasks.} 
    \label{tab:parser}    
    \vspace{-0.1cm} 
    \begin{tabular}{p{2.8cm}|p{1.3cm}|p{1.3cm}|p{1.3cm}}
            \toprule
             \specialrule{0em}{2pt}{0pt}{\ Syntax Parser}&\ \ $\bm{D}$$\mapsto $$\bm{B}$&\ \ $\bm{D}$$\mapsto $$\bm{E}$&\ \ $\bm{D}$$\mapsto $$\bm{K}$\\
            \midrule
            \emph{(1) Without Graph} & \ \ \ \  86.6&\ \ \ \ 85.9&\ \ \ \ \ 88.1\\
            \midrule
            \emph{\textbf{(2) Stanford Graph}} &\ \ \ \ \textbf{87.1}&\ \ \ \ \textbf{86.6}&\ \ \ \ \ \textbf{88.6} \\
            \emph{\ \ \ \ \ +compare with (1)} &\ \ \ \small{\textbf{(+0.5)}}&\ \ \ \small{\textbf{(+0.7)}}&\ \ \ \ \small{\textbf{(+0.5)}} \\
            \midrule
%            \emph{Without Graph} & \ \ \ \  87.1&\ \ \ \ 86.6&\ \ \ \ \ 88.2\\
%            \emph{Stanford Graph} & \ \ \ \  87.1&\ \ \ \ 86.6&\ \ \ \ \ 88.2\\
            \specialrule{0em}{3pt}{0pt}
            \emph{\textbf{(3) Biafﬁne Graph}} &\ \ \ \ \textbf{87.9}&\ \ \ \ \textbf{87.3}&\ \ \ \ \ \textbf{89.1} \\
            \emph{\ \ \ \ \ +compare with (1)} &\ \ \ \small{\textbf{(+1.3)}}&\ \ \ \small{\textbf{(+1.4)}}&\ \ \ \ \small{\textbf{(+1.0)}} \\
            \emph{\ \ \ \ \ +compare with (2)} &\ \ \ \small{\textbf{(+0.8)}}&\ \ \ \small{\textbf{(+0.7)}}&\ \ \ \ \small{\textbf{(+0.5)}} \\
            \bottomrule
    \end{tabular}
    \vspace{-0.1cm}
\end{table}

\begin{table}
\linespread{0.88} 
    \centering
            \caption{The Influence of model depth (i.e., attention heads) on $\bm{D}$$\mapsto $$\bm{*}$ tasks. The metric is accuracy (\%).} 
    \label{tab:att} 
    \vspace{-0.1cm}
    \begin{tabular}{p{2.5cm}|p{1.35cm}<{\centering}|p{1.35cm}<{\centering}|p{1.35cm}<{\centering}}
            \toprule
            Models&$\bm{D}$$\mapsto $$\bm{B}$&$\bm{D}$$\mapsto $$\bm{E}$&$\bm{D}$$\mapsto $$\bm{K}$\\
            \midrule
            \emph{HGAT w 1 head}& \ \ 86.9&\ \ 86.4&\ \ \ 87.5\\
            \emph{HGAT w 2 head} &\ \ 87.2&\ \ 86.8&\ \ \ 88.4 \\
            \emph{\textbf{HGAT w 3 head}}  &\ \ \textbf{87.9}&\ \ \textbf{87.3}&\ \ \ \textbf{89.1}\\
            \emph{HGAT w 4 head} & \ \ 87.7&\ \ 87.2&\ \ \ 88.7\\
            \emph{HGAT w 5 head} & \ \ 87.5&\ \ 86.9&\ \ \ 88.2\\
            \midrule
            \emph{Trans. w 5 head} & \ \ 86.6&\ \ 86.1&\ \ \ 88.4\\
            \emph{Trans. w 6 head} & \ \ 87.6&\ \ 86.7&\ \ \ 88.7\\
            \emph{Trans. w 7 head} & \ \ 87.5&\ \ 87.0&\ \ \ 89.1\\
            \emph{\textbf{Trans. w 8 head}} & \ \ \textbf{87.9}&\ \ \textbf{87.3}&\ \ \ \textbf{89.1}\\
            \emph{Trans. w 9 head} & \ \ 86.8&\ \ 87.1&\ \ \ 88.8\\
            \emph{Trans. w 10 head} & \ \ 87.2&\ \ 87.3&\ \ \ 89.0\\
            \bottomrule
    \end{tabular}
%    \vspace{-0.15cm}
\end{table}

\subsection{Hyper-parameter Study}
As explained in previous studies~\cite{velivckovic2018graph,vaswani2017attention}, multi-head attention can extract the relationship among words from multiple subspaces, thus it may increase the representation capability of our GAST model. We conduct experiments on the effect of multi-head attention by changing the number of attention heads. As the results shown in table~\ref{tab:att}, the best performance is achieved when we utilize three heads in HGAT and eight heads in POS-Transformer. With fewer heads, the performance drops at most 1.3\% at $\bm{D}$$\mapsto $$\bm{B}$, 1.2\% at $\bm{D}$$\mapsto $$\bm{E}$ and 1.6\% at $\bm{D}$$\mapsto $$\bm{K}$, indicating the information from some crucial subspace is lost. However, even if we stack more heads, the performance also decreases at a similar level. It is likely due to the redundant feature interactions with more heads, which may weaken the performance of GAST as well.

\subsection{Visualization of Adaptive Embedding}
To further demonstrate the transferability of GAST, we visualize the learned feature embeddings of the original Glove, BERT-DAAT, and GAST-BERT (incorporating syntactic graph embeddings). Due to space constraints, we only show one transfer task (i.e., $\bm{B}$$\mapsto $$\bm{D}$). Specifically, in subfigure (a)$\sim$(c), we sample 500 reviews from domain  $\bm{B}$ and 500 reviews from domain  $\bm{D}$. Moreover, we also sample 1,000 reviews (500 positive and 500 negative) randomly from the target domain $\bm{D}$ and project their feature embeddings via t-SNE~\cite{maaten2008visualizing}. 

\begin{figure}
	\centering
	\vspace{0.12cm}
	\includegraphics[width=0.98\columnwidth]{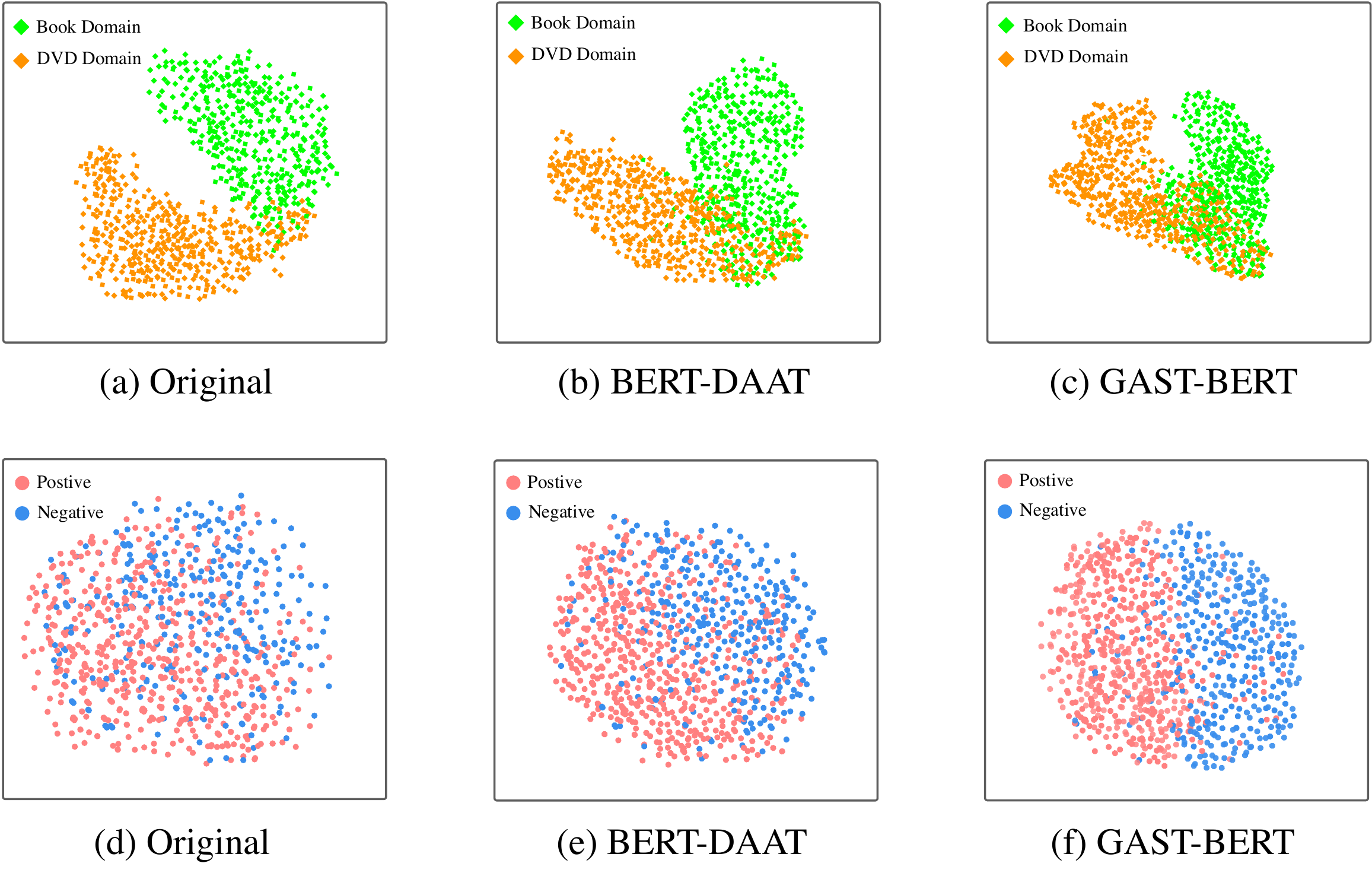}
	\vspace{-0.15cm}
	\caption{The t-SNE projection of the extracted features. The above three subfigures (i.e., (a)$\sim$(c)) show t-SNE visualization of different model's feature embedding for the $\bm{B}$$\mapsto $$\bm{D}$ task. The red and blue points in (d)$\sim$(f) denote the target positive and target negative examples, respectively.}
	\label{fig:dis}
	\vspace{-0.25cm}
\end{figure}

As shown in Figure~\ref{fig:dis}, from the original word representation (i.e., subfigure (a)) to final feature embeddings (i.e., subfigure (b) and (c)), the feature distributions between the source domain and target domains become more indistinguishable. The observation indicates that GAST-BERT can match the discrepancy between domain distribution and learn better domain-shared features than other methods (e.g., BERT-DAAT). Besides, the distribution of the original embedding is scattered and has a vague sentiment classification boundary. By contrast, the edge of GAST-BERT is more distinguishable than the original and BERT-DAAT. The reason may lie in that GAST-BERT can distill the domain-invariant and encode graphical sentiment knowledge so that the model can obtain a more discriminative sentiment feature in the target domain. 

In summary, the observations above indicate that our proposed GAST-BERT produces features that are easier to transfer across domains since it is domain awareness and helps to distill graphical feature embeddings from domain-invariant syntactic structures.

\section{Conclusion}
This paper presented a focused study on leveraging graph adaptive syntactic knowledge and sequential semantics for cross-domain sentiment classification. In particular, we proposed a novel domain adaptive method for CDSC called the Graph Adaptive Semantic Transfer (GAST) model, which mainly consists of two modules. The first is the POS-Transformer module, which is able to learn the overall semantic features from word sequences and POS tags. The second module is Hybrid Graph Attention (HGAT), designed to learn domain-invariant syntactic features from the syntactic graph. Moreover, the two modules are optimized by an integrated adaptive strategy, which ensures GAST understands invariant features better between domains. Experiments on four public datasets verified the effectiveness of our model. The ablation results and case studies further illustrate each module's point and explainability, which may provide insights for future work.

\section{Acknowledgements} 
This research was partially supported by grants from the National Key Research and Development Program of China (Grant No. 2021YF\\F0901005), the National Natural Science Foundation of China (Grants No. 61922073, U20A20229 and 62006066), the Foundation of State Key Laboratory of Cognitive Intelligence, iFLYTEK, P.R.China (No. CI0S-2020SC05) and the Open Project Program of the National Laboratory of Pattern Recognition (NLPR). 

%\newpage
\appendix
\begin{appendices}
\vspace{0.3cm}
\section{} 
\vspace{-0.25cm}
\begin{figure}[h]
	\centering
	\includegraphics[width=1\columnwidth]{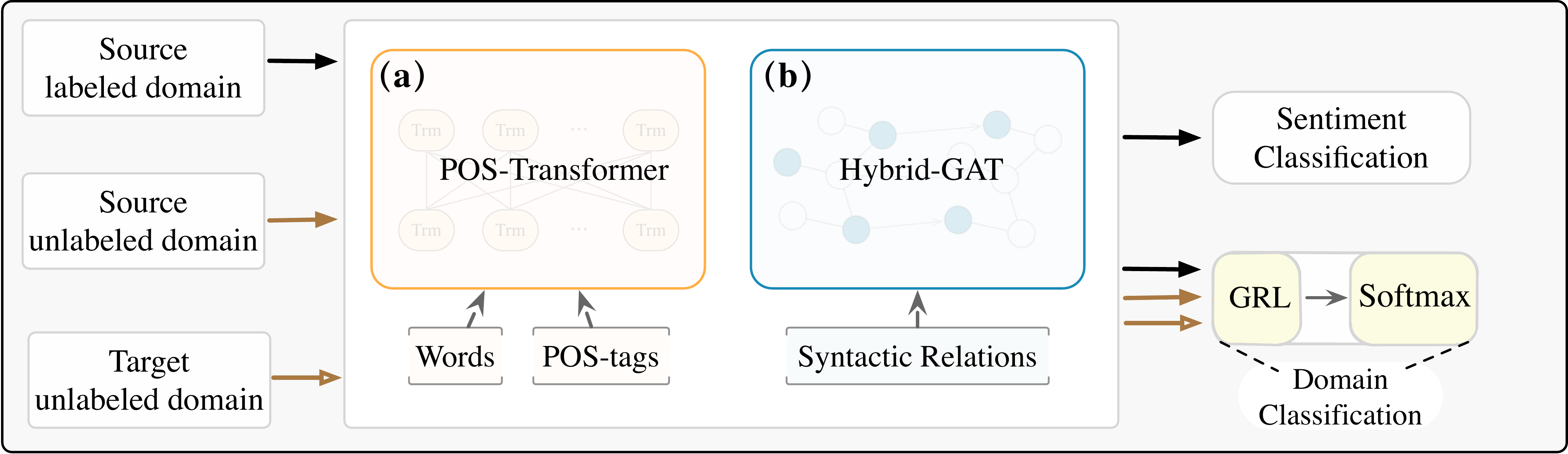}
	\vspace{-0.4cm}
	\caption{The framework of the ablation model \textbf{G\_\emph{Non\_IDS} as described in section~\ref{ab}}. It mainly includes two tasks, i.e., sentiment classification and domain classification.
	}
	\label{fig:app}
	\vspace{-0.1cm}
\end{figure}

The proposed GAST model achieves sentiment domain adaptation by IDS, which aims to facilitate knowledge transfer across domains. However, to better evaluate the performance of IDS, we replace it with a widely used adaptive strategy (i.e., Gradient Reversal Layer, GRL~\cite{ganin2016domain}). Specifically, we treat the output of HGAT (i.e., $H$) as the final feature and feed into the \emph{softmax} layer for domain classification. The goal of the domain classification task is to identify whether the example originates from source domain (i.e., ${\mathcal{D}}_s$) or target domain (i.e., ${\mathcal{D}}_t$). 
The formulation can be defined as:
\begin{equation}
	\begin{split}
	&\hat{y}^{d}\ =\ softmax({{W}_d R + {b}_d})\ .
	\end{split}
	\label{eq:y_sen}
\end{equation}

The traditional training method is to minimize the classification error of the domain classifier so that the classifier can better distinguish the difference between domains. It means that minimize the domain classification loss will enforce the domain classifier to learn domain-specific features, which is contrary to our purpose( i.e., learn domain-invariant features). As mentioned before, our goal is to learn domain-invariant features that can not be discriminated between domains and thus we need to maximize the loss of the domain classifier. However, in this way, the training purpose of the sentiment classifier (i.e., minimize the sentiment classification error) and the domain classifier (i.e., maximize the domain classification error) will compete against each other, in an adversarial way. 
To eliminate this problem, we introduce a Gradient Reversal Layer (GRL)~\cite{ganin2016domain,li2017end,li2018hierarchical,zhang2019interactive,du2020adversarial} to reverse the gradient during training progress so that both loss functions can be trained jointly. Formally, during the forward propagation, the GRL acts as an identity transformation $G(\cdot)$. During the back-propagation, the GRL takes the gradient from the subsequent level and changes its sign, i.e., multiplies it by $-1$, before passing to the preceding layer:
\begin{alignat}{2}
	&G(x)\ =\ x\ ,\ \ \ 
	&\frac{\partial G(x)}{\partial x}\ =\ - I\ .
\end{alignat}

Note that, the GRL ensures that the feature distributions over the two domains are more similar, thus resulting in the domain-invariant features. 
Through this way, the domain classifier can be trained by minimizing the \emph{cross-entropy} for all data from the source domain ${\mathcal{D}}_s$ and the target domain ${\mathcal{D}}_t$:
\begin{equation}
	L_{d}=-\frac{1}{N} \sum_{i=1}^{N}\left({y}_{d}^i \ln \hat{y}_{d}^i+\left(1-{y}_{d}^i\right) \ln \left(1-\hat{y}_{d}^i\right)\right),
\end{equation}
where $y_d^i$ is the ground-truth of sample $i$. $N$ is the sum of $n_s$  from source domain and $n_t$ from target domain.
Due to space limitations, we omit the calculation process of GRL. For details, please refer to~\cite{ganin2016domain} or related work~\cite{li2017end,li2018hierarchical,zhang2019interactive,du2020adversarial}.

 \end{appendices}

%\newpage 
%\newpage
\balance
\bibliography{sigir22_ref_cdsc.bib} \balance


\begin{thebibliography}{49}

%%% the Web address.

\ifx \showCODEN    \undefined \def \showCODEN     #1{\unskip}     \fi
\ifx \showDOI      \undefined \def \showDOI       #1{#1}\fi
\ifx \showISBNx    \undefined \def \showISBNx     #1{\unskip}     \fi
\ifx \showISBNxiii \undefined \def \showISBNxiii  #1{\unskip}     \fi
\ifx \showISSN     \undefined \def \showISSN      #1{\unskip}     \fi
\ifx \showLCCN     \undefined \def \showLCCN      #1{\unskip}     \fi
\ifx \shownote     \undefined \def \shownote      #1{#1}          \fi
\ifx \showarticletitle \undefined \def \showarticletitle #1{#1}   \fi
\ifx \showURL      \undefined \def \showURL       {\relax}        \fi
% invisible to TeX
\providecommand\bibfield[2]{#2}
\providecommand\bibinfo[2]{#2}
\providecommand\natexlab[1]{#1}
\providecommand\showeprint[2][]{arXiv:#2}

\bibitem[\protect\citeauthoryear{Blitzer, Dredze, and Pereira}{Blitzer
  et~al\mbox{.}}{2007}]%
        {blitzer2007biographies}
\bibfield{author}{\bibinfo{person}{John Blitzer}, \bibinfo{person}{Mark
  Dredze}, {and} \bibinfo{person}{Fernando Pereira}.}
  \bibinfo{year}{2007}\natexlab{}.
\newblock \showarticletitle{Biographies, bollywood, boom-boxes and blenders:
  Domain adaptation for sentiment classification}. In
  \bibinfo{booktitle}{\emph{Proceedings of the Annual Meeting of the
  Association for Computational Linguistics}}.
\newblock


\bibitem[\protect\citeauthoryear{Blitzer, McDonald, and Pereira}{Blitzer
  et~al\mbox{.}}{2006}]%
        {blitzer2006domain}
\bibfield{author}{\bibinfo{person}{John Blitzer}, \bibinfo{person}{Ryan
  McDonald}, {and} \bibinfo{person}{Fernando Pereira}.}
  \bibinfo{year}{2006}\natexlab{}.
\newblock \showarticletitle{Domain adaptation with structural correspondence
  learning}. In \bibinfo{booktitle}{\emph{Proceedings of the 2006 conference on
  empirical methods in natural language processing}}.
  \bibinfo{pages}{120--128}.
\newblock


\bibitem[\protect\citeauthoryear{Cai and Wan}{Cai and Wan}{2019}]%
        {cai2019multi}
\bibfield{author}{\bibinfo{person}{Yitao Cai} {and} \bibinfo{person}{Xiaojun
  Wan}.} \bibinfo{year}{2019}\natexlab{}.
\newblock \showarticletitle{Multi-domain sentiment classification based on
  domain-aware embedding and attention}. In
  \bibinfo{booktitle}{\emph{Proceedings of the 28th International Joint
  Conference on Artificial Intelligence}}. AAAI Press,
  \bibinfo{pages}{4904--4910}.
\newblock


\bibitem[\protect\citeauthoryear{Chen and Manning}{Chen and Manning}{2014}]%
        {chen2014fast}
\bibfield{author}{\bibinfo{person}{Danqi Chen} {and}
  \bibinfo{person}{Christopher~D Manning}.} \bibinfo{year}{2014}\natexlab{}.
\newblock \showarticletitle{A fast and accurate dependency parser using neural
  networks}. In \bibinfo{booktitle}{\emph{Proceedings of the 2014 conference on
  empirical methods in natural language processing (EMNLP)}}.
  \bibinfo{pages}{740--750}.
\newblock


\bibitem[\protect\citeauthoryear{Chen, Ma, and Xiao}{Chen
  et~al\mbox{.}}{2018}]%
        {chen2018fastgcn}
\bibfield{author}{\bibinfo{person}{Jie Chen}, \bibinfo{person}{Tengfei Ma},
  {and} \bibinfo{person}{Cao Xiao}.} \bibinfo{year}{2018}\natexlab{}.
\newblock \showarticletitle{FastGCN: Fast Learning with Graph Convolutional
  Networks via Importance Sampling}. In \bibinfo{booktitle}{\emph{International
  Conference on Learning Representations}}.
\newblock


\bibitem[\protect\citeauthoryear{Chen, Xu, Weinberger, and Sha}{Chen
  et~al\mbox{.}}{2012}]%
        {chen2012marginalized}
\bibfield{author}{\bibinfo{person}{Minmin Chen}, \bibinfo{person}{Zhixiang Xu},
  \bibinfo{person}{Kilian~Q Weinberger}, {and} \bibinfo{person}{Fei Sha}.}
  \bibinfo{year}{2012}\natexlab{}.
\newblock \showarticletitle{Marginalized denoising autoencoders for domain
  adaptation}. In \bibinfo{booktitle}{\emph{Proceedings of the 29th
  International Coference on International Conference on Machine Learning}}.
  \bibinfo{pages}{1627--1634}.
\newblock


\bibitem[\protect\citeauthoryear{Ding, Yu, and Jiang}{Ding
  et~al\mbox{.}}{2017}]%
        {ding2017recurrent}
\bibfield{author}{\bibinfo{person}{Ying Ding}, \bibinfo{person}{Jianfei Yu},
  {and} \bibinfo{person}{Jing Jiang}.} \bibinfo{year}{2017}\natexlab{}.
\newblock \showarticletitle{Recurrent neural networks with auxiliary labels for
  cross-domain opinion target extraction}. In
  \bibinfo{booktitle}{\emph{Proceedings of the AAAI Conference on Artificial
  Intelligence}}, Vol.~\bibinfo{volume}{31}.
\newblock


\bibitem[\protect\citeauthoryear{Dozat and Manning}{Dozat and Manning}{2016}]%
        {dozat2016deep}
\bibfield{author}{\bibinfo{person}{Timothy Dozat} {and}
  \bibinfo{person}{Christopher~D Manning}.} \bibinfo{year}{2016}\natexlab{}.
\newblock \showarticletitle{Deep biaffine attention for neural dependency
  parsing}.
\newblock \bibinfo{journal}{\emph{arXiv preprint arXiv:1611.01734}}
  (\bibinfo{year}{2016}).
\newblock


\bibitem[\protect\citeauthoryear{Du, Sun, Wang, Qi, and Liao}{Du
  et~al\mbox{.}}{2020}]%
        {du2020adversarial}
\bibfield{author}{\bibinfo{person}{Chunning Du}, \bibinfo{person}{Haifeng Sun},
  \bibinfo{person}{Jingyu Wang}, \bibinfo{person}{Qi Qi}, {and}
  \bibinfo{person}{Jianxin Liao}.} \bibinfo{year}{2020}\natexlab{}.
\newblock \showarticletitle{Adversarial and domain-aware bert for cross-domain
  sentiment analysis}. In \bibinfo{booktitle}{\emph{Proceedings of the Annual
  Meeting of the Association for Computational Linguistics}}.
  \bibinfo{pages}{4019--4028}.
\newblock


\bibitem[\protect\citeauthoryear{Ganin and Lempitsky}{Ganin and
  Lempitsky}{2014}]%
        {ganin2014unsupervised}
\bibfield{author}{\bibinfo{person}{Yaroslav Ganin} {and}
  \bibinfo{person}{Victor Lempitsky}.} \bibinfo{year}{2014}\natexlab{}.
\newblock \showarticletitle{Unsupervised domain adaptation by backpropagation}.
\newblock \bibinfo{journal}{\emph{arXiv preprint arXiv:1409.7495}}
  (\bibinfo{year}{2014}).
\newblock


\bibitem[\protect\citeauthoryear{Ganin, Ustinova, Ajakan, Germain, Larochelle,
  Laviolette, Marchand, and Lempitsky}{Ganin et~al\mbox{.}}{2016}]%
        {ganin2016domain}
\bibfield{author}{\bibinfo{person}{Yaroslav Ganin}, \bibinfo{person}{Evgeniya
  Ustinova}, \bibinfo{person}{Hana Ajakan}, \bibinfo{person}{Pascal Germain},
  \bibinfo{person}{Hugo Larochelle}, \bibinfo{person}{Fran{\c{c}}ois
  Laviolette}, \bibinfo{person}{Mario Marchand}, {and} \bibinfo{person}{Victor
  Lempitsky}.} \bibinfo{year}{2016}\natexlab{}.
\newblock \showarticletitle{Domain-adversarial training of neural networks}.
\newblock \bibinfo{journal}{\emph{The Journal of Machine Learning Research}}
  \bibinfo{volume}{17}, \bibinfo{number}{1} (\bibinfo{year}{2016}),
  \bibinfo{pages}{2096--2030}.
\newblock


\bibitem[\protect\citeauthoryear{Ghosal, Hazarika, Roy, Majumder, Mihalcea, and
  Poria}{Ghosal et~al\mbox{.}}{2020}]%
        {ghosal2020kingdom}
\bibfield{author}{\bibinfo{person}{Deepanway Ghosal},
  \bibinfo{person}{Devamanyu Hazarika}, \bibinfo{person}{Abhinaba Roy},
  \bibinfo{person}{Navonil Majumder}, \bibinfo{person}{Rada Mihalcea}, {and}
  \bibinfo{person}{Soujanya Poria}.} \bibinfo{year}{2020}\natexlab{}.
\newblock \showarticletitle{KinGDOM: Knowledge-Guided DOMain adaptation for
  sentiment analysis}.
\newblock \bibinfo{journal}{\emph{arXiv preprint arXiv:2005.00791}}
  (\bibinfo{year}{2020}).
\newblock


\bibitem[\protect\citeauthoryear{Glorot, Bordes, and Bengio}{Glorot
  et~al\mbox{.}}{2011}]%
        {glorot2011domain}
\bibfield{author}{\bibinfo{person}{Xavier Glorot}, \bibinfo{person}{Antoine
  Bordes}, {and} \bibinfo{person}{Yoshua Bengio}.}
  \bibinfo{year}{2011}\natexlab{}.
\newblock \showarticletitle{Domain Adaptation for Large-Scale Sentiment
  Classification: A Deep Learning Approach}. In
  \bibinfo{booktitle}{\emph{International Conference on Machine learning
  (ICML)}}. Omnipress, \bibinfo{pages}{513--520}.
\newblock


\bibitem[\protect\citeauthoryear{Gong, Yu, and Xia}{Gong et~al\mbox{.}}{2020}]%
        {gong2020unified}
\bibfield{author}{\bibinfo{person}{Chenggong Gong}, \bibinfo{person}{Jianfei
  Yu}, {and} \bibinfo{person}{Rui Xia}.} \bibinfo{year}{2020}\natexlab{}.
\newblock \showarticletitle{Unified Feature and Instance Based Domain
  Adaptation for End-to-End Aspect-based Sentiment Analysis}. In
  \bibinfo{booktitle}{\emph{Proceedings of the 2020 Conference on Empirical
  Methods in Natural Language Processing (EMNLP)}}.
  \bibinfo{pages}{7035--7045}.
\newblock


\bibitem[\protect\citeauthoryear{He, Lee, Ng, and Dahlmeier}{He
  et~al\mbox{.}}{2018}]%
        {he2018adaptive}
\bibfield{author}{\bibinfo{person}{Ruidan He}, \bibinfo{person}{Wee~Sun Lee},
  \bibinfo{person}{Hwee~Tou Ng}, {and} \bibinfo{person}{Daniel Dahlmeier}.}
  \bibinfo{year}{2018}\natexlab{}.
\newblock \showarticletitle{Adaptive Semi-supervised Learning for Cross-domain
  Sentiment Classification}. In \bibinfo{booktitle}{\emph{Proceedings of the
  Conference on Empirical Methods in Natural Language Processing}}.
\newblock


\bibitem[\protect\citeauthoryear{Hochreiter and Schmidhuber}{Hochreiter and
  Schmidhuber}{1997}]%
        {hochreiter1997long}
\bibfield{author}{\bibinfo{person}{Sepp Hochreiter} {and}
  \bibinfo{person}{J{\"u}rgen Schmidhuber}.} \bibinfo{year}{1997}\natexlab{}.
\newblock \showarticletitle{Long short-term memory}.
\newblock \bibinfo{journal}{\emph{Neural computation}} \bibinfo{volume}{9},
  \bibinfo{number}{8} (\bibinfo{year}{1997}), \bibinfo{pages}{1735--1780}.
\newblock


\bibitem[\protect\citeauthoryear{Hu and Liu}{Hu and Liu}{2004}]%
        {hu2004mining}
\bibfield{author}{\bibinfo{person}{Minqing Hu} {and} \bibinfo{person}{Bing
  Liu}.} \bibinfo{year}{2004}\natexlab{}.
\newblock \showarticletitle{Mining and summarizing customer reviews}. In
  \bibinfo{booktitle}{\emph{Proceedings of the tenth ACM SIGKDD international
  conference on Knowledge discovery and data mining}}.
  \bibinfo{pages}{168--177}.
\newblock


\bibitem[\protect\citeauthoryear{Huang and Carley}{Huang and Carley}{2019}]%
        {huang2019syntax}
\bibfield{author}{\bibinfo{person}{Binxuan Huang} {and}
  \bibinfo{person}{Kathleen~M Carley}.} \bibinfo{year}{2019}\natexlab{}.
\newblock \showarticletitle{Syntax-Aware Aspect Level Sentiment Classification
  with Graph Attention Networks}. In \bibinfo{booktitle}{\emph{Proceedings of
  the 2019 Conference on Empirical Methods in Natural Language Processing and
  the 9th International Joint Conference on Natural Language Processing}}.
  \bibinfo{pages}{5469--5477}.
\newblock


\bibitem[\protect\citeauthoryear{Kenton and Toutanova}{Kenton and
  Toutanova}{2019}]%
        {kenton2019bert}
\bibfield{author}{\bibinfo{person}{Jacob Devlin Ming-Wei~Chang Kenton} {and}
  \bibinfo{person}{Lee~Kristina Toutanova}.} \bibinfo{year}{2019}\natexlab{}.
\newblock \showarticletitle{BERT: Pre-training of Deep Bidirectional
  Transformers for Language Understanding}. In
  \bibinfo{booktitle}{\emph{Proceedings of NAACL-HLT}}.
  \bibinfo{pages}{4171--4186}.
\newblock


\bibitem[\protect\citeauthoryear{Kingma and Ba}{Kingma and Ba}{2015}]%
        {kingma2015adam}
\bibfield{author}{\bibinfo{person}{Diederik~P Kingma} {and}
  \bibinfo{person}{Jimmy Ba}.} \bibinfo{year}{2015}\natexlab{}.
\newblock \showarticletitle{Adam: A Method for Stochastic Optimization}. In
  \bibinfo{booktitle}{\emph{ICLR (Poster)}}.
\newblock


\bibitem[\protect\citeauthoryear{Lample, Conneau, Denoyer, and Ranzato}{Lample
  et~al\mbox{.}}{2018}]%
        {lample2018unsupervised}
\bibfield{author}{\bibinfo{person}{Guillaume Lample}, \bibinfo{person}{Alexis
  Conneau}, \bibinfo{person}{Ludovic Denoyer}, {and}
  \bibinfo{person}{Marc'Aurelio Ranzato}.} \bibinfo{year}{2018}\natexlab{}.
\newblock \showarticletitle{Unsupervised Machine Translation Using Monolingual
  Corpora Only}. In \bibinfo{booktitle}{\emph{International Conference on
  Learning Representations}}.
\newblock


\bibitem[\protect\citeauthoryear{Li, Pan, Jin, Yang, and Zhu}{Li
  et~al\mbox{.}}{2012}]%
        {li2012cross}
\bibfield{author}{\bibinfo{person}{Fangtao Li}, \bibinfo{person}{Sinno~Jialin
  Pan}, \bibinfo{person}{Ou Jin}, \bibinfo{person}{Qiang Yang}, {and}
  \bibinfo{person}{Xiaoyan Zhu}.} \bibinfo{year}{2012}\natexlab{}.
\newblock \showarticletitle{Cross-domain co-extraction of sentiment and topic
  lexicons}. In \bibinfo{booktitle}{\emph{Proceedings of the 50th Annual
  Meeting of the Association for Computational Linguistics (Volume 1: Long
  Papers)}}. \bibinfo{pages}{410--419}.
\newblock


\bibitem[\protect\citeauthoryear{Li, Ye, Long, Tang, Xu, and Wang}{Li
  et~al\mbox{.}}{2020b}]%
        {li2020simultaneous}
\bibfield{author}{\bibinfo{person}{Liang Li}, \bibinfo{person}{Weirui Ye},
  \bibinfo{person}{Mingsheng Long}, \bibinfo{person}{Yateng Tang},
  \bibinfo{person}{Jin Xu}, {and} \bibinfo{person}{Jianmin Wang}.}
  \bibinfo{year}{2020}\natexlab{b}.
\newblock \showarticletitle{Simultaneous learning of pivots and representations
  for cross-domain sentiment classification}. In
  \bibinfo{booktitle}{\emph{Proceedings of the AAAI Conference on Artificial
  Intelligence}}, Vol.~\bibinfo{volume}{34}. \bibinfo{pages}{8220--8227}.
\newblock


\bibitem[\protect\citeauthoryear{Li, Chen, Zhang, Dong, and Keutzer}{Li
  et~al\mbox{.}}{2020a}]%
        {li2020cross}
\bibfield{author}{\bibinfo{person}{Tian Li}, \bibinfo{person}{Xiang Chen},
  \bibinfo{person}{Shanghang Zhang}, \bibinfo{person}{Zhen Dong}, {and}
  \bibinfo{person}{Kurt Keutzer}.} \bibinfo{year}{2020}\natexlab{a}.
\newblock \showarticletitle{Cross-Domain Sentiment Classification with
  In-Domain Contrastive Learning}.
\newblock \bibinfo{journal}{\emph{arXiv preprint arXiv:2012.02943}}
  (\bibinfo{year}{2020}).
\newblock


\bibitem[\protect\citeauthoryear{Li, Wei, Zhang, and Yang}{Li
  et~al\mbox{.}}{2018}]%
        {li2018hierarchical}
\bibfield{author}{\bibinfo{person}{Zheng Li}, \bibinfo{person}{Ying Wei},
  \bibinfo{person}{Yu Zhang}, {and} \bibinfo{person}{Qiang Yang}.}
  \bibinfo{year}{2018}\natexlab{}.
\newblock \showarticletitle{Hierarchical attention transfer network for
  cross-domain sentiment classification}. In
  \bibinfo{booktitle}{\emph{Thirty-Second AAAI Conference on Artificial
  Intelligence}}.
\newblock


\bibitem[\protect\citeauthoryear{Li, Zhang, Wei, Wu, and Yang}{Li
  et~al\mbox{.}}{2017}]%
        {li2017end}
\bibfield{author}{\bibinfo{person}{Zheng Li}, \bibinfo{person}{Yun Zhang},
  \bibinfo{person}{Ying Wei}, \bibinfo{person}{Yuxiang Wu}, {and}
  \bibinfo{person}{Qiang Yang}.} \bibinfo{year}{2017}\natexlab{}.
\newblock \showarticletitle{End-to-End Adversarial Memory Network for
  Cross-domain Sentiment Classification.}. In
  \bibinfo{booktitle}{\emph{IJCAI}}. \bibinfo{pages}{2237--2243}.
\newblock


\bibitem[\protect\citeauthoryear{Liu}{Liu}{2012}]%
        {liu2012sentiment}
\bibfield{author}{\bibinfo{person}{Bing Liu}.} \bibinfo{year}{2012}\natexlab{}.
\newblock \showarticletitle{Sentiment analysis and opinion mining}.
\newblock \bibinfo{journal}{\emph{Synthesis lectures on human language
  technologies}} \bibinfo{volume}{5}, \bibinfo{number}{1}
  (\bibinfo{year}{2012}), \bibinfo{pages}{1--167}.
\newblock


\bibitem[\protect\citeauthoryear{Long, Cao, Wang, and Jordan}{Long
  et~al\mbox{.}}{2015}]%
        {long2015learning}
\bibfield{author}{\bibinfo{person}{Mingsheng Long}, \bibinfo{person}{Yue Cao},
  \bibinfo{person}{Jianmin Wang}, {and} \bibinfo{person}{Michael Jordan}.}
  \bibinfo{year}{2015}\natexlab{}.
\newblock \showarticletitle{Learning transferable features with deep adaptation
  networks}. In \bibinfo{booktitle}{\emph{International conference on machine
  learning}}. PMLR, \bibinfo{pages}{97--105}.
\newblock


\bibitem[\protect\citeauthoryear{Maaten and Hinton}{Maaten and Hinton}{2008}]%
        {maaten2008visualizing}
\bibfield{author}{\bibinfo{person}{Laurens van~der Maaten} {and}
  \bibinfo{person}{Geoffrey Hinton}.} \bibinfo{year}{2008}\natexlab{}.
\newblock \showarticletitle{Visualizing data using t-SNE}.
\newblock \bibinfo{journal}{\emph{Journal of machine learning research}}
  \bibinfo{volume}{9}, \bibinfo{number}{Nov} (\bibinfo{year}{2008}),
  \bibinfo{pages}{2579--2605}.
\newblock


\bibitem[\protect\citeauthoryear{Mahalakshmi and Sivasankar}{Mahalakshmi and
  Sivasankar}{2015}]%
        {mahalakshmi2015cross}
\bibfield{author}{\bibinfo{person}{S Mahalakshmi} {and} \bibinfo{person}{E
  Sivasankar}.} \bibinfo{year}{2015}\natexlab{}.
\newblock \showarticletitle{Cross domain sentiment analysis using different
  machine learning techniques}. In \bibinfo{booktitle}{\emph{Proceedings of the
  Fifth International Conference on Fuzzy and Neuro Computing (FANCCO-2015)}}.
  Springer, \bibinfo{pages}{77--87}.
\newblock


\bibitem[\protect\citeauthoryear{Meijer, Frasincar, and
  Tru{\c{s}}c{\u{a}}}{Meijer et~al\mbox{.}}{2021}]%
        {meijer2021explaining}
\bibfield{author}{\bibinfo{person}{Lisa Meijer}, \bibinfo{person}{Flavius
  Frasincar}, {and} \bibinfo{person}{Maria~Mihaela Tru{\c{s}}c{\u{a}}}.}
  \bibinfo{year}{2021}\natexlab{}.
\newblock \showarticletitle{Explaining a neural attention model for
  aspect-based sentiment classification using diagnostic classification}. In
  \bibinfo{booktitle}{\emph{Proceedings of the 36th Annual ACM Symposium on
  Applied Computing}}. \bibinfo{pages}{821--827}.
\newblock


\bibitem[\protect\citeauthoryear{Pan, Ni, Sun, Yang, and Chen}{Pan
  et~al\mbox{.}}{2010}]%
        {pan2010cross}
\bibfield{author}{\bibinfo{person}{Sinno~Jialin Pan},
  \bibinfo{person}{Xiaochuan Ni}, \bibinfo{person}{Jian-Tao Sun},
  \bibinfo{person}{Qiang Yang}, {and} \bibinfo{person}{Zheng Chen}.}
  \bibinfo{year}{2010}\natexlab{}.
\newblock \showarticletitle{Cross-domain sentiment classification via spectral
  feature alignment}. In \bibinfo{booktitle}{\emph{Proceedings of the 19th
  international conference on World wide web}}. \bibinfo{pages}{751--760}.
\newblock


\bibitem[\protect\citeauthoryear{Pang, Lee, and Vaithyanathan}{Pang
  et~al\mbox{.}}{2002}]%
        {pang2002thumbs}
\bibfield{author}{\bibinfo{person}{Bo Pang}, \bibinfo{person}{Lillian Lee},
  {and} \bibinfo{person}{Shivakumar Vaithyanathan}.}
  \bibinfo{year}{2002}\natexlab{}.
\newblock \showarticletitle{Thumbs up?: sentiment classification using machine
  learning techniques}. In \bibinfo{booktitle}{\emph{Proceedings of the ACL-02
  conference on Empirical methods in natural language processing-Volume 10}}.
  Association for Computational Linguistics, \bibinfo{pages}{79--86}.
\newblock


\bibitem[\protect\citeauthoryear{Pennington, Socher, and Manning}{Pennington
  et~al\mbox{.}}{2014}]%
        {pennington2014glove}
\bibfield{author}{\bibinfo{person}{Jeffrey Pennington},
  \bibinfo{person}{Richard Socher}, {and} \bibinfo{person}{Christopher~D
  Manning}.} \bibinfo{year}{2014}\natexlab{}.
\newblock \showarticletitle{Glove: Global vectors for word representation}. In
  \bibinfo{booktitle}{\emph{Proceedings of the 2014 conference on empirical
  methods in natural language processing (EMNLP)}}.
  \bibinfo{pages}{1532--1543}.
\newblock


\bibitem[\protect\citeauthoryear{Shi, Han, Song, Wang, Wang, Du, and
  Philip}{Shi et~al\mbox{.}}{2019}]%
        {shi2019deep}
\bibfield{author}{\bibinfo{person}{Chuan Shi}, \bibinfo{person}{Xiaotian Han},
  \bibinfo{person}{Li Song}, \bibinfo{person}{Xiao Wang},
  \bibinfo{person}{Senzhang Wang}, \bibinfo{person}{Junping Du}, {and}
  \bibinfo{person}{S~Yu Philip}.} \bibinfo{year}{2019}\natexlab{}.
\newblock \showarticletitle{Deep collaborative filtering with multi-aspect
  information in heterogeneous networks}.
\newblock \bibinfo{journal}{\emph{IEEE transactions on knowledge and data
  engineering}} \bibinfo{volume}{33}, \bibinfo{number}{4}
  (\bibinfo{year}{2019}), \bibinfo{pages}{1413--1425}.
\newblock


\bibitem[\protect\citeauthoryear{Sun, Zhang, Mensah, Mao, and Liu}{Sun
  et~al\mbox{.}}{2019}]%
        {sun2019aspect}
\bibfield{author}{\bibinfo{person}{Kai Sun}, \bibinfo{person}{Richong Zhang},
  \bibinfo{person}{Samuel Mensah}, \bibinfo{person}{Yongyi Mao}, {and}
  \bibinfo{person}{Xudong Liu}.} \bibinfo{year}{2019}\natexlab{}.
\newblock \showarticletitle{Aspect-level sentiment analysis via convolution
  over dependency tree}. In \bibinfo{booktitle}{\emph{Proceedings of the 2019
  Conference on Empirical Methods in Natural Language Processing and the 9th
  International Joint Conference on Natural Language Processing
  (EMNLP-IJCNLP)}}. \bibinfo{pages}{5683--5692}.
\newblock


\bibitem[\protect\citeauthoryear{Tian, Chen, and Song}{Tian
  et~al\mbox{.}}{2021}]%
        {tian2021aspect}
\bibfield{author}{\bibinfo{person}{Yuanhe Tian}, \bibinfo{person}{Guimin Chen},
  {and} \bibinfo{person}{Yan Song}.} \bibinfo{year}{2021}\natexlab{}.
\newblock \showarticletitle{Aspect-based Sentiment Analysis with Type-aware
  Graph Convolutional Networks and Layer Ensemble}. In
  \bibinfo{booktitle}{\emph{Proceedings of the 2021 Conference of the North
  American Chapter of the Association for Computational Linguistics: Human
  Language Technologies}}. \bibinfo{pages}{2910--2922}.
\newblock


\bibitem[\protect\citeauthoryear{Vaswani, Shazeer, Parmar, Uszkoreit, Jones,
  Gomez, Kaiser, and Polosukhin}{Vaswani et~al\mbox{.}}{2017}]%
        {vaswani2017attention}
\bibfield{author}{\bibinfo{person}{Ashish Vaswani}, \bibinfo{person}{Noam
  Shazeer}, \bibinfo{person}{Niki Parmar}, \bibinfo{person}{Jakob Uszkoreit},
  \bibinfo{person}{Llion Jones}, \bibinfo{person}{Aidan~N Gomez},
  \bibinfo{person}{{\L}ukasz Kaiser}, {and} \bibinfo{person}{Illia
  Polosukhin}.} \bibinfo{year}{2017}\natexlab{}.
\newblock \showarticletitle{Attention is all you need}. In
  \bibinfo{booktitle}{\emph{Advances in neural information processing
  systems}}. \bibinfo{pages}{5998--6008}.
\newblock


\bibitem[\protect\citeauthoryear{Veli{\v{c}}kovi{\'c}, Cucurull, Casanova,
  Romero, Li{\`o}, and Bengio}{Veli{\v{c}}kovi{\'c} et~al\mbox{.}}{2018}]%
        {velivckovic2018graph}
\bibfield{author}{\bibinfo{person}{Petar Veli{\v{c}}kovi{\'c}},
  \bibinfo{person}{Guillem Cucurull}, \bibinfo{person}{Arantxa Casanova},
  \bibinfo{person}{Adriana Romero}, \bibinfo{person}{Pietro Li{\`o}}, {and}
  \bibinfo{person}{Yoshua Bengio}.} \bibinfo{year}{2018}\natexlab{}.
\newblock \showarticletitle{Graph Attention Networks}. In
  \bibinfo{booktitle}{\emph{International Conference on Learning
  Representations}}.
\newblock


\bibitem[\protect\citeauthoryear{Wang, Shen, Yang, Quan, and Wang}{Wang
  et~al\mbox{.}}{2020}]%
        {wang2020relational}
\bibfield{author}{\bibinfo{person}{Kai Wang}, \bibinfo{person}{Weizhou Shen},
  \bibinfo{person}{Yunyi Yang}, \bibinfo{person}{Xiaojun Quan}, {and}
  \bibinfo{person}{Rui Wang}.} \bibinfo{year}{2020}\natexlab{}.
\newblock \showarticletitle{Relational Graph Attention Network for Aspect-based
  Sentiment Analysis}. In \bibinfo{booktitle}{\emph{Proceedings of the 58th
  Annual Meeting of the Association for Computational Linguistics}}.
\newblock


\bibitem[\protect\citeauthoryear{Wang and Pan}{Wang and Pan}{2018}]%
        {wang2018recursive}
\bibfield{author}{\bibinfo{person}{Wenya Wang} {and}
  \bibinfo{person}{Sinno~Jialin Pan}.} \bibinfo{year}{2018}\natexlab{}.
\newblock \showarticletitle{Recursive neural structural correspondence network
  for cross-domain aspect and opinion co-extraction}. In
  \bibinfo{booktitle}{\emph{Proceedings of the 56th Annual Meeting of the
  Association for Computational Linguistics (Volume 1: Long Papers)}}.
  \bibinfo{pages}{2171--2181}.
\newblock


\bibitem[\protect\citeauthoryear{Xia and Zong}{Xia and Zong}{2011}]%
        {xia2011pos}
\bibfield{author}{\bibinfo{person}{Rui Xia} {and} \bibinfo{person}{Chengqing
  Zong}.} \bibinfo{year}{2011}\natexlab{}.
\newblock \showarticletitle{A POS-based ensemble model for cross-domain
  sentiment classification}. In \bibinfo{booktitle}{\emph{Proceedings of 5th
  international joint conference on natural language processing}}.
  \bibinfo{pages}{614--622}.
\newblock


\bibitem[\protect\citeauthoryear{Yao, Mao, and Luo}{Yao et~al\mbox{.}}{2019}]%
        {yao2019graph}
\bibfield{author}{\bibinfo{person}{Liang Yao}, \bibinfo{person}{Chengsheng
  Mao}, {and} \bibinfo{person}{Yuan Luo}.} \bibinfo{year}{2019}\natexlab{}.
\newblock \showarticletitle{Graph convolutional networks for text
  classification}. In \bibinfo{booktitle}{\emph{Proceedings of the AAAI
  Conference on Artificial Intelligence}}, Vol.~\bibinfo{volume}{33}.
  \bibinfo{pages}{7370--7377}.
\newblock


\bibitem[\protect\citeauthoryear{Yu and Jiang}{Yu and Jiang}{2016}]%
        {yu-jiang-2016-learning}
\bibfield{author}{\bibinfo{person}{Jianfei Yu} {and} \bibinfo{person}{Jing
  Jiang}.} \bibinfo{year}{2016}\natexlab{}.
\newblock \showarticletitle{Learning Sentence Embeddings with Auxiliary Tasks
  for Cross-Domain Sentiment Classification}. In
  \bibinfo{booktitle}{\emph{Proceedings of the 2016 Conference on Empirical
  Methods in Natural Language Processing}}. \bibinfo{pages}{236--246}.
\newblock


\bibitem[\protect\citeauthoryear{Yuan, Zhao, Qin, and Liu}{Yuan
  et~al\mbox{.}}{2021}]%
        {yuan2021learning}
\bibfield{author}{\bibinfo{person}{Jianhua Yuan}, \bibinfo{person}{Yanyan
  Zhao}, \bibinfo{person}{Bing Qin}, {and} \bibinfo{person}{Ting Liu}.}
  \bibinfo{year}{2021}\natexlab{}.
\newblock \showarticletitle{Learning to Share by Masking the Non-shared for
  Multi-domain Sentiment Classification}.
\newblock \bibinfo{journal}{\emph{arXiv preprint arXiv:2104.08480}}
  (\bibinfo{year}{2021}).
\newblock


\bibitem[\protect\citeauthoryear{Zhang, Liu, Qian, Xiang, Cui, Zhou, and
  Chen}{Zhang et~al\mbox{.}}{2021}]%
        {zhang2021eatn}
\bibfield{author}{\bibinfo{person}{Kai Zhang}, \bibinfo{person}{Qi Liu},
  \bibinfo{person}{Hao Qian}, \bibinfo{person}{Biao Xiang},
  \bibinfo{person}{Qing Cui}, \bibinfo{person}{Jun Zhou}, {and}
  \bibinfo{person}{Enhong Chen}.} \bibinfo{year}{2021}\natexlab{}.
\newblock \showarticletitle{EATN: An Efficient Adaptive Transfer Network for
  Aspect-level Sentiment Analysis}.
\newblock \bibinfo{journal}{\emph{IEEE Transactions on Knowledge and Data
  Engineering}} (\bibinfo{year}{2021}).
\newblock


\bibitem[\protect\citeauthoryear{Zhang, Zhang, Liu, Zhao, Zhu, and Chen}{Zhang
  et~al\mbox{.}}{2019}]%
        {zhang2019interactive}
\bibfield{author}{\bibinfo{person}{Kai Zhang}, \bibinfo{person}{Hefu Zhang},
  \bibinfo{person}{Qi Liu}, \bibinfo{person}{Hongke Zhao},
  \bibinfo{person}{Hengshu Zhu}, {and} \bibinfo{person}{Enhong Chen}.}
  \bibinfo{year}{2019}\natexlab{}.
\newblock \showarticletitle{Interactive attention transfer network for
  cross-domain sentiment classification}. In
  \bibinfo{booktitle}{\emph{Proceedings of the AAAI Conference on Artificial
  Intelligence}}.
\newblock


\bibitem[\protect\citeauthoryear{Zhang, Zhang, Zhang, Zhao, Liu, Wu, and
  Chen}{Zhang et~al\mbox{.}}{2022}]%
        {zhang2022incorporating}
\bibfield{author}{\bibinfo{person}{Kai Zhang}, \bibinfo{person}{Kun Zhang},
  \bibinfo{person}{Mengdi Zhang}, \bibinfo{person}{Hongke Zhao},
  \bibinfo{person}{Qi Liu}, \bibinfo{person}{Wei Wu}, {and}
  \bibinfo{person}{Enhong Chen}.} \bibinfo{year}{2022}\natexlab{}.
\newblock \showarticletitle{Incorporating Dynamic Semantics into Pre-Trained
  Language Model for Aspect-based Sentiment Analysis}.
\newblock \bibinfo{journal}{\emph{arXiv preprint:2203.16369}}
  (\bibinfo{year}{2022}).
\newblock


\bibitem[\protect\citeauthoryear{Zou, Tang, Xie, and Liu}{Zou
  et~al\mbox{.}}{2015}]%
        {zou2015sentiment}
\bibfield{author}{\bibinfo{person}{Huang Zou}, \bibinfo{person}{Xinhua Tang},
  \bibinfo{person}{Bin Xie}, {and} \bibinfo{person}{Bing Liu}.}
  \bibinfo{year}{2015}\natexlab{}.
\newblock \showarticletitle{Sentiment classification using machine learning
  techniques with syntax features}. In \bibinfo{booktitle}{\emph{2015
  International Conference on Computational Science and Computational
  Intelligence}}. \bibinfo{pages}{175--179}.
\newblock


\end{thebibliography}

\end{document}